\documentclass[journal]{IEEEtran}
\usepackage{graphicx}
\usepackage{url}
\usepackage{epsfig}
\usepackage{cite}
\usepackage{graphicx}
\usepackage{amsmath,amsthm,graphicx,booktabs,amsfonts,stfloats,multirow}
\usepackage{amssymb}
\usepackage{enumerate}
\usepackage{subcaption}
\usepackage{array}
\usepackage{algorithm}
\usepackage{float}
\usepackage{array}
\usepackage{epstopdf}
\usepackage{bm}
\usepackage{color}
\usepackage{latexsym}
\usepackage[table,xcdraw]{xcolor}
\usepackage{lineno}
\usepackage{colortbl}
\usepackage{arydshln}
\usepackage{multirow}
\usepackage{multicol}
\usepackage{bigstrut}

\usepackage{algpseudocode}
\def\ie{{\textit{i.e.}}}
\def\eg{{\textit{e.g.}}}

\def\etal{{\textit{et al.~}}}

\def\x{{\mathbf x}}
\def\y{{\mathbf y}}
\def\p{{\mathbf p}}
\def\a{{\mathbf a}}
\def\f{{\mathbf f}}
\def\g{{\mathbf g}}
\def\u{{\mathbf u}}

\def\A{{\mathbf A}}
\def\I{{\mathbf I}}

\def\M{{\mathbf M}}

\def\I{{\mathbf I}}
\def\U{{\mathbf U}}

\def\X{{\mathbf X}}
\def\Y{{\mathbf Y}}
\def\bfLambda{{\mathbf \Lambda}}

\def\RR{{\mathbb R}}
\def\cN{{\mathcal N}}

\begin{document}

\title{Graph Signal Processing for Geometric Data and Beyond: Theory and Applications}

\author{Wei~Hu,~\IEEEmembership{Senior Member,~IEEE,}
	    Jiahao~Pang,~\IEEEmembership{Member,~IEEE,}
	    Xianming~Liu,~\IEEEmembership{Member,~IEEE,}
	    Dong~Tian,~\IEEEmembership{Senior~Member,~IEEE,}
	    Chia-Wen~Lin,~\IEEEmembership{Fellow,~IEEE,}
	    and Anthony~Vetro,~\IEEEmembership{Fellow,~IEEE}

        \thanks{Manuscript received March 31, 2021. {\it (Corresponding author: Chia-Wen Lin)}}
        \thanks{Wei Hu is with Wangxuan Institute of Computer Technology, Peking University, Beijing, China. (e-mail: forhuwei@pku.edu.cn)}
        \thanks{Jiahao Pang and Dong Tian are with InterDigital, Princeton, NJ, USA. (e-mail: jiahao.pang@interdigital.com, dong.tian@interdigital.com)}
        \thanks{Xianming Liu is with Harbin Institute of Technology, Harbin, China. (e-mail: csxm@hit.edu.cn)}
        \thanks{Chia-Wen Lin is with Department of Electrical Engineering and Institute of Communications Engineering, National Tsing Hua University, Hsinchu, Taiwan, and with Electronic and Optoelectronic System Research Laboratories, Industrial Technology Research Institute. (e-mail: cwlin@ee.nthu.edu.tw)}
        \thanks{Anthony Vetro is with Mitsubishi Electric Research Laboratories, Cambridge, MA, USA. (e-mail: avetro@merl.com)}
}

\markboth{IEEE TMM Overview Article}%
{Shell \MakeLowercase{\textit{et al.}}: Bare Demo of IEEEtran.cls for IEEE Journals}

\maketitle

\newcommand{\red}[1] {\textcolor[rgb]{1.0,0.0,0.0}{{#1}}}
\newcommand{\blue}[1] {\textcolor[rgb]{0.0,0.0,1.0}{{#1}}}

\begin{abstract}

Geometric data acquired from real-world scenes, \eg, 2D depth images, 3D point clouds, and 4D dynamic point clouds, have found a wide range of applications including immersive telepresence, autonomous driving, surveillance, {\it etc}. 
Due to irregular sampling patterns of most geometric data, traditional image/video processing methodologies are limited, while Graph Signal Processing (GSP)---a fast-developing field in the signal processing community---enables processing signals that reside on irregular domains and plays a critical role in numerous applications of geometric data from low-level processing to high-level analysis. 
To further advance the research in this field, we provide the first timely and comprehensive overview of GSP methodologies for geometric data in a unified manner by bridging the connections between geometric data and graphs, among the various geometric data modalities, and with spectral/nodal graph filtering techniques. 
We also discuss the recently developed Graph Neural Networks (GNNs) and interpret the operation of these networks from the perspective of GSP. 
We conclude with a brief discussion of open problems and challenges.

\end{abstract}

% Note that keywords are not normally used for peer review papers.

\IEEEpeerreviewmaketitle

\vspace{-0.05in}
\section{Introduction}
\label{sec:intro}
\IEEEPARstart{R}{ecent} advances in depth sensing, laser scanning and image processing have enabled convenient acquisition and extraction of geometric data from real-world scenes, which can be digitized and formatted in a number of different ways. Efficiently representing, processing, and analyzing geometric data is central to a wide range of applications from augmented and virtual reality \cite{burdea2003virtual,schmalstieg2016augmented} to autonomous driving \cite{ChenLFVW:20} and surveillance/monitoring applications~\cite{benedek20143d}.

\begin{figure*}
    \centering
    \includegraphics[width=0.7\textwidth]{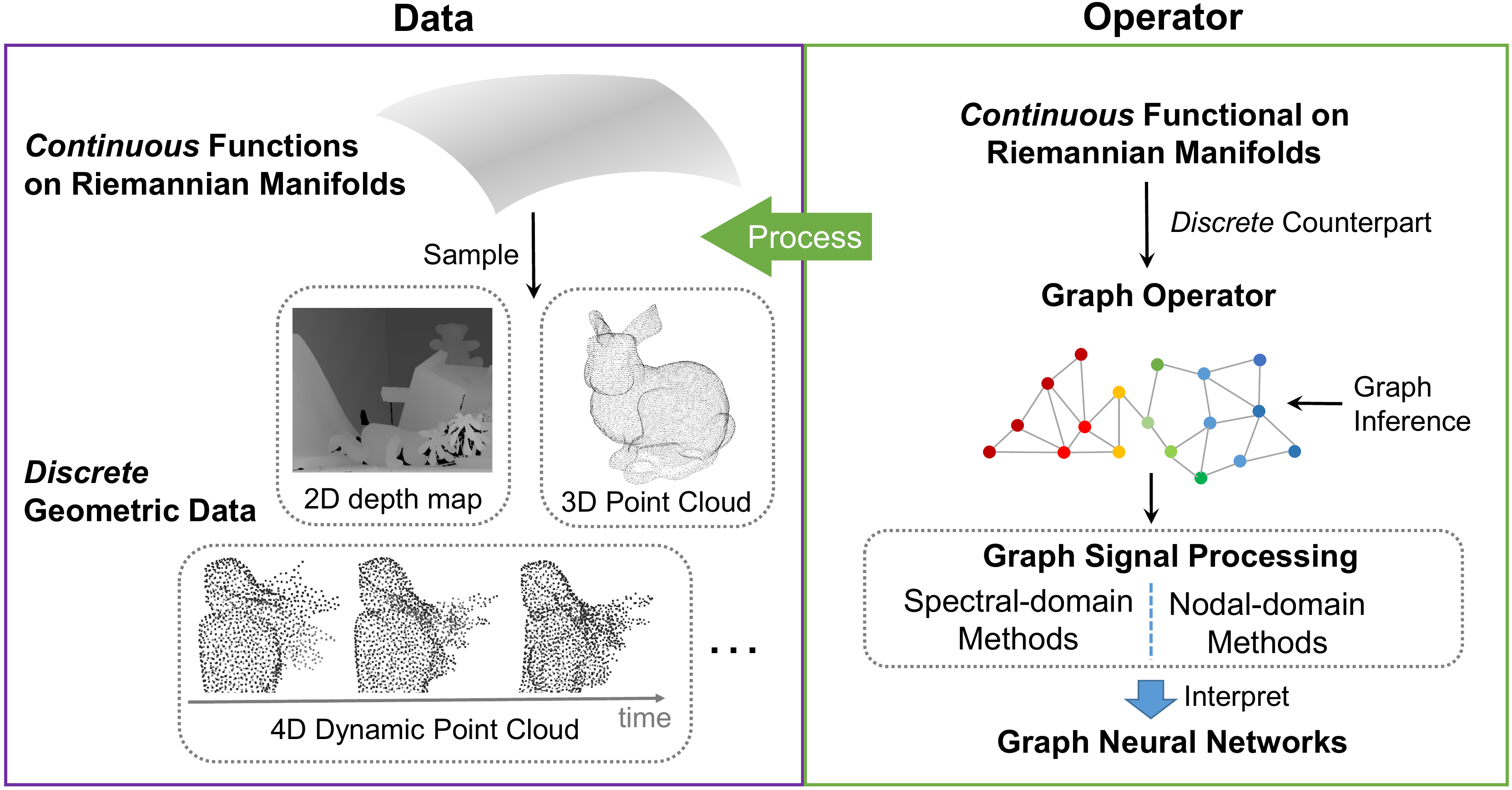}
    \caption{Illustration of GSP for geometric data processing.}
    \label{fig:framework}
\end{figure*}

Geometric data may be represented in various data formats. 
It has been recognized by Adelson, {\it et al}.~\cite{adelson91pleno} that different representations of a scene can be expressed as approximations of the plenoptic function, which is a high-dimensional mathematical representation that provides complete information about any point within a scene and also how it changes when observed from different positions. This connection among the different scene representations has also been embraced and reflected in the work plans for the development of the JPEG Pleno standardization framework~\cite{ebrahimi2016pleno}. 
In this paper, we mainly consider {\it explicit} representations of geometry, which directly describe the underlying geometry, but the framework and techniques extend to {\it implicit} representations of geometry, in which the underlying geometry is present in the data but needs to be inferred, \eg, from camera data. 
Examples of explicit geometric representations include 2D geometric data (\eg, depth maps), 3D geometric data (\eg, point clouds and meshes), and 4D geometric data (\eg, dynamic point clouds), as demonstrated in Fig.~\ref{fig:framework}. Examples of implicit geometric representations include camera-based inputs, \eg, multiview video. For many cases of interest that aim to render immersive imagery of a scene, the focus will be on dense representations of geometry. However, there are also some applications of interest that benefit from sparse representations of geometry, such as human activity analysis, in which the geometry of the human body can be represented with few data points. 

Traditional image/video processing techniques assume sampling patterns over regular grids and have limitations when dealing with the wide range of geometric data formats, some of which have irregular sampling patterns. 
To overcome the limitations of traditional techniques, Graph Signal Processing (GSP) techniques have been proposed and developed in recent years to process signals that reside over connected graph nodes~\cite{shuman2013emerging,sandryhaila2013discrete,ortega2018graph}. 
For geometric data, each sample is denoted by a graph node and the associated 3D coordinate (or depth) is the signal to be analyzed. The underlying surface of geometric data provides an intrinsic graph connectivity or graph topology.   
The graph-based representation has several advantages over conventional representations in that it is more \emph{compact} and \emph{accurate}, and \emph{structure-adaptive} since it naturally captures geometric characteristics in the data, such as piece-wise smoothness (PWS)~\cite{hu2014multiresolution}.

A unified framework of GSP for geometric data is illustrated in Fig.~\ref{fig:framework}, in which we highlight how geometric data and graph operators are counterparts in the context of Riemannian manifolds.
Given continuous functions on Riemannian manifolds, geometric data are discrete samples of the functions representing the geometry of objects, which often lies on a low-dimensional manifold, \eg, 3D point clouds essentially represent 2D surfaces embedded in the 3D space. 
Correspondingly, graph operators are discrete counterparts of the continuous functionals defined on Riemannian manifolds.
Theoretically, it has been shown that graph operators converge to functionals on Riemannian manifolds under certain constraints~\cite{ting2010analysis}, 
while graph regularizers converge to smooth functionals on Riemannian manifolds that is capable of enforcing low dimensionality of data~\cite{hein2006uniform,pang2017,zeng20183d}.
Hence, GSP tools are naturally advantageous for geometric data processing by representing the underlying topology of geometry on graphs. 

A graph operator is typically constructed based on domain knowledge or inferred from training data as shown in Fig.~\ref{fig:framework}.
It essentially specifies a \emph{graph filtering} process, which can be performed either in the spectral-domain (\ie, graph transform domain)~\cite{hammond2011wavelets} or nodal-domain (\ie, spatial domain)~\cite{gadde2013bilateral}, which are referred to as {\it spectral-domain GSP methods} and {\it nodal-domain GSP methods}, respectively.
Nodal-domain methods typically avoid eigen-decomposition for fast computing over large-scale data while still relying on spectral analysis to provide insights~\cite{tian2014chebyshev}.
A nodal-domain method might also be specified through a graph regularizer to enforce graph-signal smoothness~\cite{zeng20183d,hu2019feature}. 
%rossi2018geometry
\emph{Sparsity} and \emph{smoothness} are two widely used domain models. 
Additionally, Graph Neural Networks (GNNs) have been developed to enable inference with graph signals including geometric data \cite{bronstein2017geometric}, which are often motivated or interpretable by GSP tools. 
Hence, methodologically, we will first elaborate on spectral-domain and nodal-domain GSP methods for geometric data respectively, then discuss the interpretability of GNNs from the perspective of GSP.

In practice, GSP for geometric data plays a critical role in numerous applications of geometric data, from low-level processing, such as restoration and compression, to high-level analysis. 
The processing of geometric data includes denoising, enhancement and resampling, as well as compression such as point cloud coding standardized in MPEG\footnote{https://mpeg.chiariglione.org/standards/mpeg-i/point-cloud-compression} and JPEG Pleno\footnote{https://jpeg.org/jpegpleno/}, while the analysis of geometric data addresses supervised or unsupervised feature learning for classification, segmentation, detection, and generation. 
These applications are unique relative to the use of GSP techniques for other data in terms of the signal model and processing methods.

This overview paper distinguishes itself from relevant review papers such as  \cite{ortega2018graph,bronstein2017geometric,wu2020comprehensive,guo2020deep,stankovic2020graph} in the following aspects. 
While \cite{ortega2018graph} provides a general overview for GSP covering core ideas in GSP and recent advances in developing basic GSP tools with a variety of applications, our paper is dedicated to GSP for geometric data with unique signal characteristics that have led to new insights and understanding.
Compared with \cite{bronstein2017geometric} and \cite{wu2020comprehensive} which provide a comprehensive overview of geometric deep learning including GNNs, we focus on those GNNs that are motivated or interpretable by GSP tools.
In comparison with \cite{guo2020deep} that reviews recent progress in deep learning methods for point clouds, we emphasize on GNNs for geometric data that are explainable via GSP, while in \cite{guo2020deep}, graph-based methods are discussed only as one of many types of approaches for 3D shape classification and point cloud segmentation without further discussion of the model interpretability. 
Furthermore, compared with \cite{stankovic2020graph} that analyzes machine learning on graphs from the {\it graph diffusion perspective} and connects different learning algorithms on graphs with different diffusion models, we emphasize the {\it graph signal processing} aspect of graph neural networks, and endeavor to interpret their behavior in both the spectral and the nodal domains, as well as several aspects to understand the representation learning of graph neural networks from the perspective of GSP as discussed in Section VI.
In summary, this paper provides an overview of GSP methods specifically for a unique and important class of data---geometric data, as well as insights into the interpretability of GNNs from the perspective of GSP tools.

The remainder of this paper is organized as follows. Section~\ref{sec:background} reviews basic concepts in GSP, graph Fourier Transform, as well as interpretation of graph variation operators both in the discrete domain and continuous domain. 
Section~\ref{sec:representation} introduces the graph representation of geometric data based on their characteristics, along with problems and applications of geometric data to be discussed throughout the paper.  
Then, we elaborate on spectral-domain GSP methods for geometric data in Section~\ref{sec:spectral} and nodal-domain GSP methods in Section~\ref{sec:nodal}. 
Next, we provide the interpretations of GNNs for geometric data from the perspective of GSP in Section~\ref{sec:gnn}. 
Finally, future directions and conclusions are discussed in Section~\ref{sec:future} and Section~\ref{sec:conclusion}, respectively.

\vspace{-0.05in}
\section{Review: Graph Signal Processing}
\label{sec:background}
\subsection{Graph Variation Operators and Graph Signal}
\label{subsec:laplacian}

We denote a graph $ \mathcal{G}=\{\mathcal{V},\mathcal{E}, \mathbf{A}\} $, which is composed of a vertex set $ \mathcal{V} $ of cardinality $|\mathcal{V}|=N$, an edge set $ \mathcal{E} $ connecting vertices, and an adjacency matrix $\mathbf{A}$. 
Each entry $a_{i,j}$ in $\mathbf{A}$ represents the weight of the edge between vertices $i$ and $j$, which often captures the similarity between adjacent vertices. 
In geometric data processing, we often consider an undirected graph with non-negative edge weights, \ie, $a_{i,j} = a_{j,i} \geq 0$.

Among variation operators in GSP, we focus on the commonly used graph Laplacian matrix. The \textit{combinatorial graph Laplacian} \cite{shuman2013emerging} is defined as $ \mathbf{L}:=\mathbf{D}-\mathbf{A} $, where $ \mathbf{D} $ is the \textit{degree matrix}---a diagonal matrix where $ d_{i,i} = \sum_{j=1}^N a_{i,j} $. 
Given real and non-negative edge weights in an undirected graph, $\mathbf{L}$ is real, symmetric, and positive semi-definite \cite{chung1997spectral}. 
The symmetrically normalized version is  $\mathcal{L}_{\text{sym}}:=\mathbf{D}^{-\frac{1}{2}}\mathbf{L}\mathbf{D}^{-\frac{1}{2}}$, and the random walk graph Laplacian is $\mathcal{L}_{\text{rw}}:=\mathbf{D}^{-1}\mathbf{L}$, which are often used for theoretical analysis or in neural networks due to the normalization property. 

A graph signal is a function that assigns a scalar or vector to each vertex. 
For simplicity, we consider $x:\mathcal{V} \rightarrow \mathbb{R}$, such as the intensity on each vertex of a mesh. 
We denote graph signals as $\x \in \mathbb{R}^{N}$, where $x_i$ represents the signal value at the $i$-th vertex. 

\begin{figure*}[htbp]
    \centering
    \includegraphics[width=0.8\textwidth]{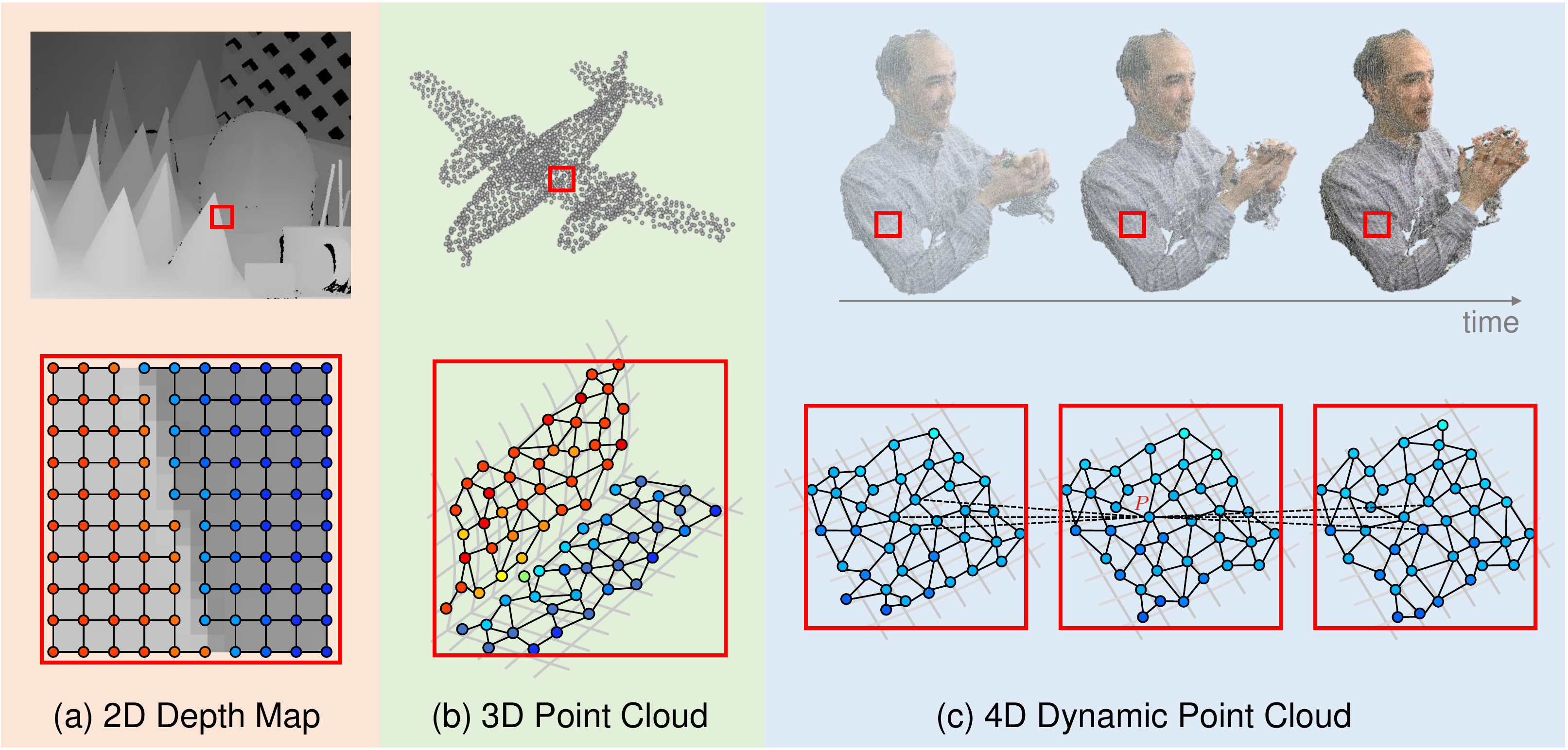}
%    \vspace{-0.1in}
    \caption{Geometric data and their graph representations. The graphs of the patches enclosed in red squares are shown at the bottom; the vertices are colored by the corresponding graph signals. (a)~2D Depth map~\cite{middlebury}. (b)~3D Point cloud~\cite{dai2017scannet}. (c)~4D dynamic point cloud \cite{Cai16}, where the temporal edges of a point $P$ are also shown.}
    \label{fig:gsp_geo}
\end{figure*}

\vspace{-0.1in}
\subsection{Graph Fourier Transform}
\label{subsec:GFT}
Because $\mathbf L$ is a real symmetric matrix, it admits an eigen-decomposition $\mathbf{L} = \U \bfLambda \U^{\top}$, where $\U=[\u_1,...,\u_N]$ is an orthonormal matrix containing the eigenvectors $\u_i$, and $\bfLambda = \mathrm{diag}(\lambda_1,...,\lambda_N)$ consists of eigenvalues $\{\lambda_1=0 \leq \lambda_2 \leq ... \leq \lambda_N\}$. 
We refer to the eigenvalue $\lambda_i$ as the {\it graph frequency/spectrum}, with a smaller eigenvalue corresponding to a lower graph frequency.

For any graph signal $\x \in \mathbb{R}^{N}$ residing on the vertices of $\mathcal G$, its graph Fourier transform (GFT) $\hat{\x}  \in \mathbb{R}^{N}$ is defined as \cite{hammond2011wavelets} 
\begin{equation}
    \hat{\x} = \U^{\top} \x. 
\end{equation}
The inverse GFT follows as 
\begin{equation}
    \x = \U \hat{\x}. 
\end{equation}
With an appropriately constructed graph that captures the signal structure well, the GFT will lead to a compact representation of the graph signal in the spectral domain, which is beneficial for geometric data processing such as reconstruction and compression.

\vspace{-0.1in}
\subsection{Interpretation of Graph Variation Operators}
\label{ssec:con_inter}
The graph variation operators have various interpretations, both in the discrete domain and the continuous domain. 

In the discrete domain, we can interpret graph Laplacian matrices by precision matrices under Gaussian-Markov Random Fields (GMRFs)~\blue{\cite{rue2005gaussian}} from a probabilistic perspective, and thus further show the GFT approximates the Karhunen-Lo\`{e}ve transform (KLT) for signal decorrelation under GMRFs. 
As discussed in \cite{egilmez2017graph}, there is a one-to-one correspondence between precision matrices of different classes of GMRFs and types of graph Laplacian matrices. 
For instance, the combinatorial graph Laplacian corresponds to the precision matrix of an attractive, DC-intrinsic GMRF. 
Further, as the eigenvectors of the precision matrix (the inverse of the covariance matrix) constitute the basis of the KLT, the GFT approximates the KLT under a family of statistical processes, as proved in different ways in \cite{zhang2012analyzing,hu2014multiresolution,hu2015intra,zhang2016graph}. 
This indicates the GFT is approximately the optimal linear transform for signal decorrelation, which is beneficial to the compression of geometric data as will be discussed in Section~\ref{subsubsec:compression}. 

In the continuous domain, instead of viewing a neighborhood graph as inherently discrete, it can be treated as a discrete approximation of a \emph{Riemannian manifold} \cite{singer2006graph,ting2010analysis}.
Thus, as the number of vertices on a graph increases, the graph is converging to a Riemannian manifold.
In this scenario, each observation of a graph signal is a discrete sample of a continuous signal (function) defined on the manifold.
Note that not all graph signals can be interpreted in the continuous domain: voting pattern in a social network or paper information in a citation network is inherently discrete. 
With a focus on geometric data which are indeed signals captured from a continuous surface, we have a continuous-domain interpretation of graph signals as discrete samples of a continuous function (Fig.\,\ref{fig:framework}).
The link between neighborhood graphs and Riemannian manifolds enables us to process geometric data with tools from differential geometry and variational methods \cite{hein2007graph}.
For instance, the graph Laplacian operator converges to the Laplace-Beltrami operator in the continuous manifold when the number of samples tends to infinity. 
Hence, without direct access to the underlying geometry (surface), it is still possible to infer the property of the geometry based on its discrete samples.

\begin{table*}[htbp]
  \centering\scriptsize
  \caption{Representative Geometric Datasets and Relevant Application Scenarios.}
    \begin{tabular}{l||l|l|l}
    \hline
    \textbf{Geometric Data Format} & \textbf{Datasets} & \textbf{Contents} & \textbf{Typical Applications/Tasks} \bigstrut\\
    \hline
    \hline
    \multirow{4}[6]{*}{2D depth map} & FlyingThings3D \cite{FlyingThings3D} & Synthetic scene & \multirow{4}[6]{*}{Stereo matching, depth completion} \bigstrut\\
\cdashline{2-3}          & Middlebury  \cite{middlebury} & \multirow{2}[2]{*}{Indoor scene} &  \bigstrut[t]\\
          & Tsukuba  \cite{Tsukuba} &       &  \bigstrut[b]\\
\cdashline{2-3}          & KITTI  \cite{geiger2012we} & Driving scene &  \bigstrut\\
    \hline
    \multirow{10}[10]{*}{3D point cloud} & Stanford 3D Scanning Repository  \cite{levoy2017stanford} & \multirow{2}[2]{*}{Single object} & \multirow{4}[4]{*}{3D telepresence, surface reconstruction} \bigstrut[t]\\
          & Benchmark  \cite{berger2013benchmark} &       &  \bigstrut[b]\\
\cdashline{2-3}          & MPEG Sequences  \cite{Eugene2017dataset} & \multirow{2}[2]{*}{Single person} &  \bigstrut[t]\\
          & Microsoft Sequences  \cite{Cai16} &       &  \bigstrut[b]\\
\cline{2-4}          & ShapeNet  \cite{chang2015shapenet} & \multirow{2}[2]{*}{Single object} & \multirow{2}[2]{*}{Classification, part segmentation} \bigstrut[t]\\
          & ModelNet  \cite{wu20153d} &       &  \bigstrut[b]\\
\cline{2-4}          & Stanford Large-Scale 3D Indoor Spaces Dataset  \cite{armeni20163d} & \multirow{2}[2]{*}{Indoor scene} & \multirow{4}[4]{*}{Semantic/instance segmentation} \bigstrut[t]\\
          & ScanNet  \cite{dai2017scannet} &       &  \bigstrut[b]\\
\cdashline{2-3}          & KITTI  \cite{geiger2012we} & \multirow{2}[2]{*}{Driving scene} &  \bigstrut[t]\\
          & WAYMO Open Dataset \cite{Waymo} &       &  \bigstrut[b]\\
    \hline
    \multirow{4}[4]{*}{4D dynamic point cloud} & MPEG Sequences  \cite{Eugene2017dataset} & \multirow{2}[2]{*}{Single person} & \multirow{2}[2]{*}{3D telepresence, compression} \bigstrut[t]\\
          & Microsoft sequences  \cite{Cai16} &       &  \bigstrut[b]\\
\cline{2-4}          & KITTI  \cite{geiger2012we} & \multirow{2}[2]{*}{Driving scene} & \multirow{2}[2]{*}{Semantic/instance segmentation, detection} \bigstrut[t]\\
          & Semantic KITTI  \cite{behley2019iccv} &       &  \bigstrut[b]\\
    \hline
    \end{tabular}%
  \label{tab:datasets}%
\end{table*}%

For a clearer presentation, Table~\ref{tab:notation} summarizes the most important mathematical symbols used in this paper.

\begin{table}[htbp]
  \centering\footnotesize
  \caption{Key notations employed in this review article.}
    \begin{tabular}{c|l}
    \hline
    Notation & \multicolumn{1}{c}{Description} \\
    \hline\hline
    $\mathcal{G}$     & The graph being studied. \\
    $\mathbf{A}$     & Graph adjacency matrix. \\
    $\mathbf{L}$     & Graph Laplacian matrix. \\
    $\mathbf{U}$     & Inverse graph Fourier transform matrix. \\
    $\mathbf{x}$     & Geometric data (graph signal) being studied. \\
    $a_{i,j}$     & Graph weight connecting vertices $i$ and $j$. \\

    $\lambda_i$     & Graph frequency/spectrum. \\
    $\hat{h}(\cdot)$     & Spectral-domain filter coefficient. \\
    $h_k$    & Nodal-domain filter coefficient. \\
    \hline
    \end{tabular}%
  \label{tab:notation}%
\end{table}%

\vspace{-0.05in}
\section{Graph Representations of Geometric Data}
\label{sec:representation}
In this section, we elaborate on the graph representations of geometric data, which arise from the unique characteristics of geometric data and serve as the basis of GSP for geometric data processing. 
Also, we discuss and compare with non-graph representations, which helps understand the advantages and insights of graph representations.

\vspace{-0.1in}
\subsection{Problems and Challenges of Geometric Data}
There exist various problems associated with geometric data, {\it e.g.}, noise, holes (incomplete data), compression artifacts, large data size, and irregular sampling. 
For instance, due to inherent limitations in the sensing capabilities and viewpoints that are acquired, geometric data often suffer from noise and holes, which will affect the subsequent rendering or downstream inference tasks since the underlying structures are deformed. 

These problems must be accounted for in the diverse range of applications that rely on geometric data, including processing (\eg, restoration and enhancement), compression, and analysis (\eg, classification, segmentation, and recognition).
Some of the representative geometric datasets along with the corresponding application scenarios are summarized in Table~\ref{tab:datasets}. 

We assert that the chosen representation of geometric data is critically important in addressing these problems and applications. 
Next, we discuss the characteristics of geometric data, which lay the foundation for using graphs for representation. 

\vspace{-0.1in}
\subsection{Characteristics of Geometric Data}
\label{subsec:characteristics}

Geometric data represent the geometric structure underlying the surface of objects and scenes in the 3D world and have unique characteristics that capture structural properties. 

For example, 2D depth maps characterize the per-pixel physical distance between objects in a 3D scene and the sensor, which usually consists of sharp boundaries and smooth interior surfaces---referred to as piece-wise smoothness (PWS) \cite{hu2014multiresolution}, as shown in Fig.~\ref{fig:gsp_geo}(a).  
The PWS property is suitable to be described by a graph, where most edge weights are $1$ for smooth surfaces and a few weights are $0$ for discontinuities across sharp boundaries. 
Such a graph construction will lead to a compact representation in the GFT domain, where most energy is concentrated on low-frequency components for the description of smooth surfaces \cite{hu2014multiresolution}.

3D geometric data such as point clouds form omnidirectional representations of a geometric structure in the 3D world. 
As shown in Fig.~\ref{fig:gsp_geo}(b), the underlying surface of the 3D geometric data often exhibits the PWS property, as given by the normals of the data~\cite{dinesh2020point}. 
Moreover, 3D point clouds lie on a 2D manifold, as they represent 2D surfaces embedded in the 3D space. 

For 4D geometric data such as dynamic point clouds, consistency/redundancy exists along the temporal dimension~\cite{thanou2016graph,fu20icme}, as shown in Fig.~\ref{fig:gsp_geo}(c). 
However, in contrast to conventional video data, the temporal correspondences in dynamic point clouds are difficult to track, mainly because 
1) the sampling pattern may vary from frame to frame due to the irregular sampling; and 
2) the number of points in each frame of a dynamic point cloud sequence may vary significantly.

Thanks to the unique signal characteristics of geometric data, we may design methods tailored for geometric data instead of methods for general graph data. 
For instance, particular graph smoothness priors may be taken into account so that methods are optimized for the PWS property of depth maps, which leads to more robust or efficient processing as will be discussed in Section~\ref{subsec:nodal_optimization}.

\vspace{-0.1in}
\subsection{Non-Graph Representations of Geometric Data}
\label{subsec:non_graph}

There exist a variety of non-graph representations of geometric data. 
For instance, depth maps are represented as gray-scale images, while 3D point clouds are often quantized onto regular voxel grids~\cite{schnabel2006octree} or projected onto a set of depth images from multiple viewpoints~\cite{Chen_2017_CVPR}. 
These quantization-based representations transform geometric data into regular Euclidean space, which is amenable to existing methods for Euclidean data such as images, videos, and regular voxel grids.

Further, implicit functions (\eg, Signed Distance Function (SDF)) for 3D shape representation have been proposed \cite{curless1996volumetric,park2019deepsdf}, which represent a shape's surface by a discrete or continuous volumetric field: the magnitude of a point in the field represents the distance to the surface boundary and the sign indicates whether the region is inside (-) or outside (+) of the shape. 
This enables the high-quality representation of the shape surface, interpolation and completion from partial and noisy 3D input data. 
Besides, the sparse tensor representation is employed due to its expressiveness and generalizability for high-dimensional spaces \cite{choy20194d}. It also allows homogeneous data representation within traditional neural network libraries as most of them support sparse tensors.

In spite of the advantages, these non-graph representations may have the following limitations: 
1) Most importantly, representing geometric data without graphs is often deficient in capturing the underlying geometric structure explicitly. 
2) Quantization-based representations are sometimes inaccurate, \eg, due to quantization loss introduced by voxelization or projection errors when a point cloud is represented by a set of images or discretized SDF; and 
3) The representations can be redundant, e.g., a voxel-based representation of point clouds still needs to represent an unoccupied space with zeros, leading to redundant storage or processing.

\vspace{-0.1in}
\subsection{Graph Representations of Geometric Data}\label{ssec:graph_rep}
In contrast, graphs provide {\it structure-adaptive}, {\it accurate}, and {\it compact} representations for geometric data, which further inspire new insights and understanding.  

To represent geometric data on a graph $\mathcal{G}=\{\mathcal{V},\mathcal{E}, \mathbf{A}\}$, we consider points in the data (\eg, pixels in depth maps, points in point clouds and meshes) as vertices $\mathcal{V}$ with cardinality $N$. 
Further, for the $i$-th point, we represent the coordinate and possibly associated attribute $(\p_i, \a_i)$ of each point as the graph signal on each vertex, where $\p_i \in \RR^2$ or $\p_i \in \RR^3$ represents the 2D or 3D coordinate of the $i$-th point (\eg, 2D for depth maps, and 3D for point clouds), and $\a_i$ represents associated attributes, such as depth values, RGB colors, reflection intensities, and surface normals. 
To ease mathematical computation, we denote the graph signal of all vertices by a matrix,
\begin{eqnarray}
  \label{eq:raw_point_representation}
  \X  \ = \  \begin{bmatrix} \x_1^T \\ \x_2^T \\ \vdots \\
    \x_N^T
  \end{bmatrix} \ \in \ \RR^{N \times d},
\end{eqnarray}
where the $i$-th row vector $\x_i^T=[\p_i^T ~ \a_i^T] \in \RR^{1 \times d}$
represents the graph signal on the $i$-th vertex and $d$ denotes the dimension of the graph signal. 

To capture the underlying structure, we use edges $\mathcal{E}$ in the graph to describe the pairwise relationship (spatio-temporal relationship for 4D geometric data \cite{gao2019optimized,fu21tmm,xu2019predictive}) between points, which is encoded in the adjacency matrix $\A$ as reviewed in Section~\ref{sec:background}. 
The construction of $\A$, \ie, graph construction, is crucial to characterize the underlying topology of geometric data. 
We classify existing graph construction methods mainly into two families: 
1) model-based graph construction, which builds graphs with models from domain knowledge \cite{shen2010edge,zhang2014point}; and
2) learning-based graph construction, which infers/learns the underlying graph from geometric data \cite{Dong19,mateos2019connecting,franceschi2019learning, jin2020graph}.  

Model-based graph construction for geometric data often assumes edge weights are inversely proportional to the affinities in coordinates, such as a $K$-nearest-neighbor graph ($K$-NN graph) and an $\epsilon$-neighborhood graph ($\epsilon$-N graph).  
A $K$-NN graph is a graph in which two vertices are connected by an edge, when their Euclidean distance is among the $K$-th smallest Euclidean distances from one point to the others; while in an $\epsilon$-N graph, two vertices are connected if their Euclidean distance is smaller than a given threshold $\epsilon$. 
A $K$-NN graph intends to maintain a constant vertex degree in the graph, which may lead to a more stable algorithm implementation; while an $\epsilon$-N graph intends to make the vertex degree reflect local point density, leading to more physical interpretation.
Though these graphs exhibit manifold convergence properties \cite{ting2010analysis,hein2007graph}, it still remains challenging to find an efficient estimation of the sparsification parameters such as $K$ and $\epsilon$ given finite and non-uniformly sampled data.    

In learning-based graph construction, the underlying graph topology is inferred or optimized from geometric data in terms of a certain optimization criterion, such as enforcing low-frequency representations of observed signals. 
For example, given a single or partial observation, \cite{hu2019feature} optimizes a distance metric from relevant feature vectors on vertices by minimizing the graph Laplacian regularizer, leading to learned edge weights.  
Besides, edge weights could be trainable in an end-to-end learning manner~\cite{li2018adaptive}. Also, general graph learning methodologies can apply to the graph construction of geometric data \cite{Dong19,mateos2019connecting,franceschi2019learning, jin2020graph}. 

\begin{figure*}[htbp]
    \centering
    \includegraphics[width=0.98\textwidth]{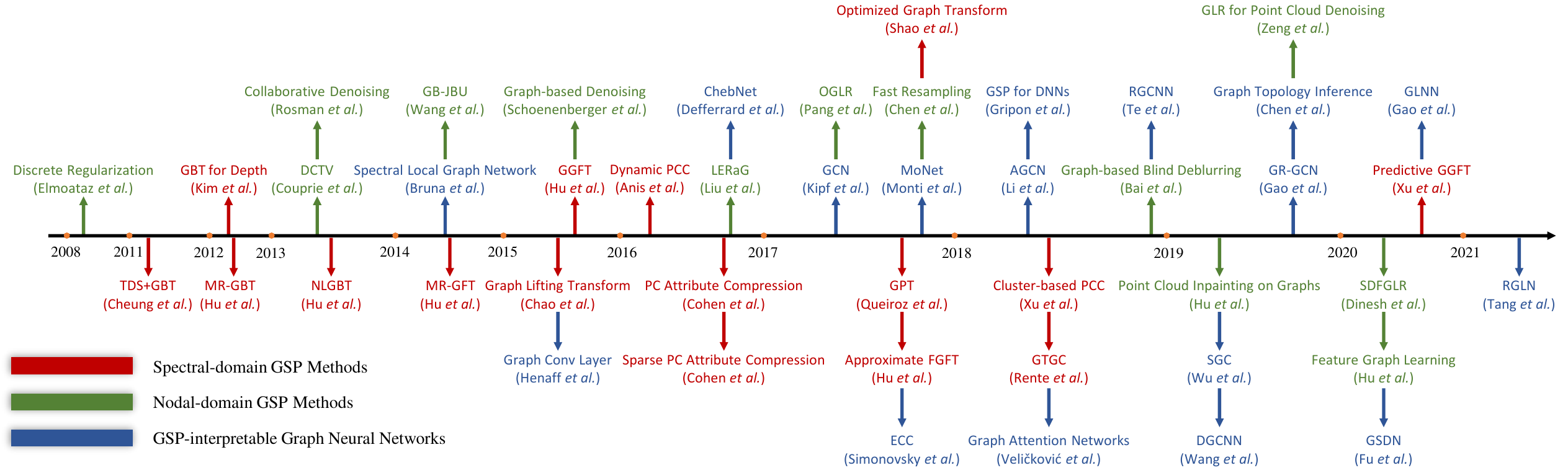}
    \caption{Representative works leveraging graph signal processing (GSP) to process or analyze geometric data.}
    \label{fig:paper_axis}
\end{figure*}

\vspace{-0.05in}
\section{Spectral-Domain GSP Methods for \\ Geometric Data}
\label{sec:spectral}
Based on the aforementioned graph representations, we will elaborate on GSP methodologies for geometric data, including spectral-domain GSP methods, nodal-domain GSP methods, and GSP-interpretable graph neural networks. 
Representative methods using GSP to process/analyze geometric data are summarized in chronological order in Fig.~\ref{fig:paper_axis}.
We start from the spectral-domain methods that offer spectral interpretations.

\vspace{-0.1in}
\subsection{Basic Principles}
Spectral-domain methods represent geometric data in the graph transform domain and perform filtering on the resulting transform coefficients. 
While various graph transforms exist, we focus our discussion on the Graph Fourier Transform (GFT) discussed in Section~\ref{subsec:GFT} without loss of generality.

Let the frequency response of a graph spectral filtering be denoted by $\hat{h}(\lambda_k)\ (k = 1,\dots, N)$, then the graph spectral filtering takes the form
\begin{eqnarray}
\Y & = & \U \begin{bmatrix}
\hat{h}(\lambda_1) & & \\
& \ddots & \\
&& \hat{h}(\lambda_{N})
\end{bmatrix}
\U^\top \X.
\label{eq:GFT_filtering_matrix}
\end{eqnarray}
This filtering first transforms the geometric data $\X$ into the GFT domain $\U^{\top}\X$, performs filtering on each eigenvalue ({\it i.e.}, the spectrum of the graph), and finally projects back to the spatial domain via the inverse GFT to acquire the filtered output $\Y$. 

As discussed in Section~\ref{ssec:con_inter}, the GFT leads to compact representations of geometric data if the constructed graph captures the underlying topology well. 
Based on the GFT representation, the key issue is to specify $N$ graph frequency responses $\{\hat{h}(\lambda_k)\}_{k=1}^N$ to operate on the geometric data; these filters should be designed according to the specific task. 
Widely used filters include {\it low-pass} graph spectral filters and {\it high-pass} graph spectral filters, which will be discussed further in the next subsection. 

Due to the computational complexity of graph transforms, which often involve full eigen-decomposition, this class of methods are either dedicated to \textit{small-scale} geometric data or applied in a \textit{divide-and-conquer} manner. 
For instance, one may divide a point cloud into regular cubes, and perform graph spectral filtering on individual cubes separately. 
Also, one may deploy a fast algorithm of GFT ({\it e.g.}, the fast GFT in \cite{le2017approximate}), to accelerate the spectral filtering process. 

\vspace{-0.1in}
\subsection{Representative Graph Spectral Filtering}\label{ssec:gs_filter}
\subsubsection{Low-Pass Graph Spectral Filtering}
\label{sssec:spectral_low_pass}
Analogous to processing digital images in the regular 2D grid, we can use a low-pass graph filter to capture the rough shape of geometric data and attenuate noise under the assumption that signals are smooth in the associated data domain. 
In practice, a geometric signal ({\it e.g.}, coordinates, normals) is inherently smooth with respect to the underlying graph, where high-frequency components are likely to be generated by fine details or noise.  
Hence, we can perform geometric data smoothing via a low-pass graph filter, essentially leading to a smoothed representation in the underlying manifold.  

One intuitive realization is an ideal low-pass graph filter, which completely eliminates all graph frequencies above a given bandwidth while keeping those below unchanged. 
The graph frequency response of an ideal low-pass graph filter with bandwidth $b$ is
\begin{equation}
    \hat{h}(\lambda_k)=\left\{
         \begin{array}{lr} 
         1,& k \leq b, \\
         0,& k > b, 
         \end{array}
     \right.
     \label{eq:low_pass_thresholding}
\end{equation}
which projects the input geometric data into a bandlimited subspace by removing components corresponding to large eigenvalues ({\it i.e.}, high-frequency components). 

\begin{figure*}[t]
  \begin{center}
    \begin{tabular}{ccccc}
    \includegraphics[width=0.17\textwidth]{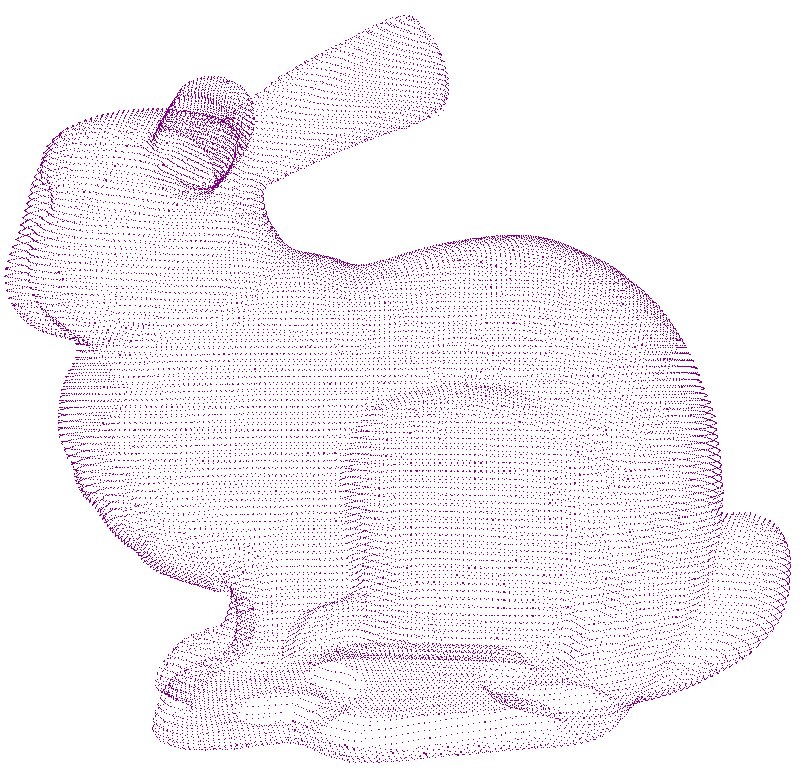} & \includegraphics[width=0.17\textwidth]{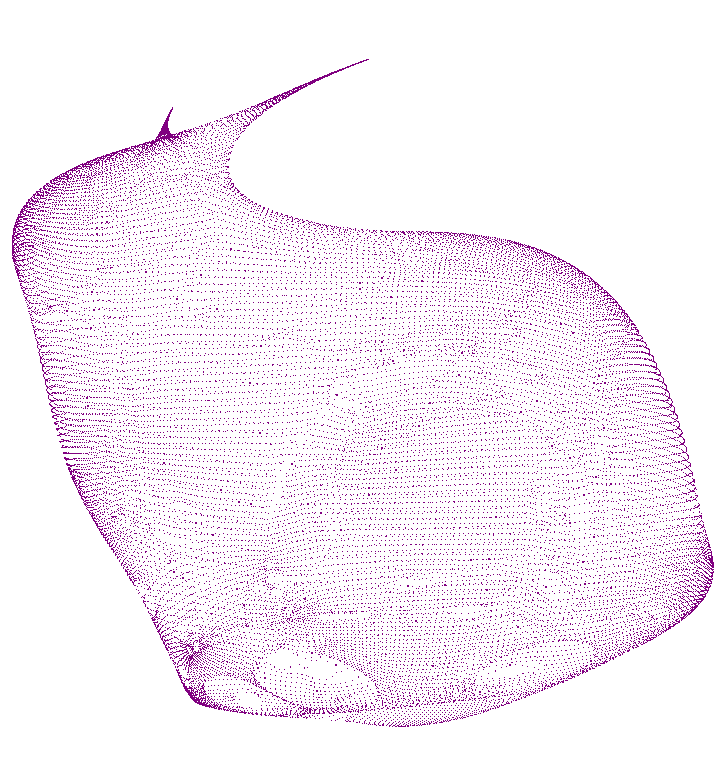} & \includegraphics[width=0.17\textwidth]{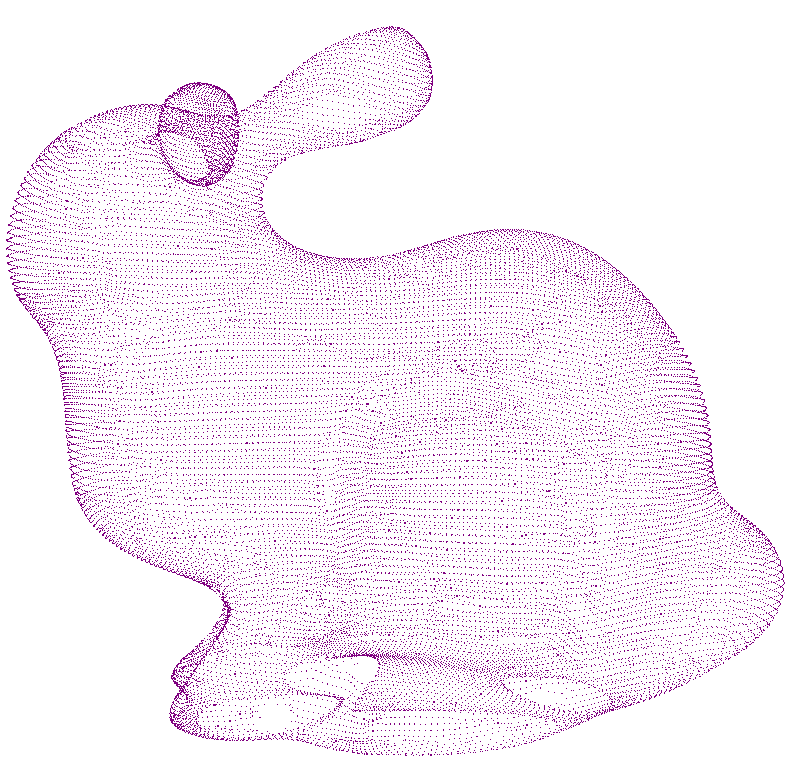} & 
    \includegraphics[width=0.17\textwidth]{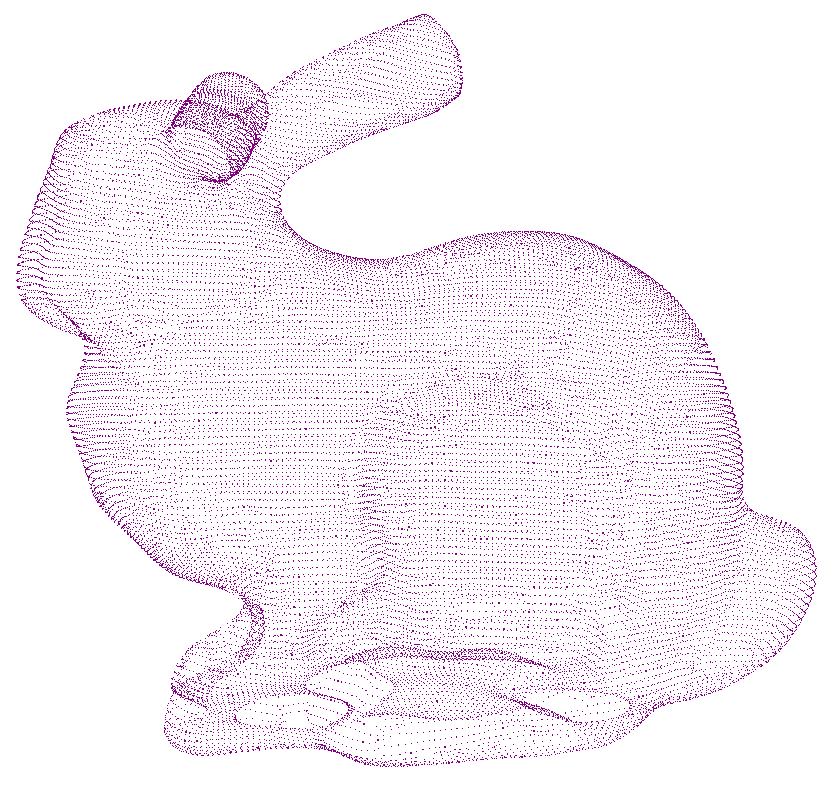} & 
    \includegraphics[width=0.2\textwidth]{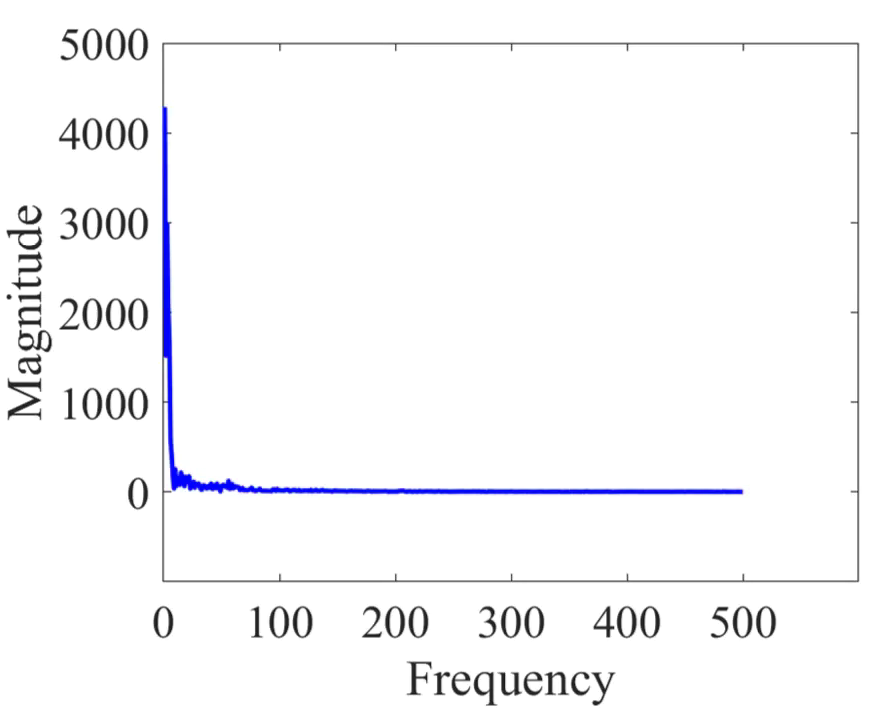}
    \\
    {\small (a) Original.} & {\small (b) 10 frequencies.} &  {\small (c) 100 frequencies. } & {\small (d) 400 frequencies.} & {\small (e) Graph spectral distribution.}  
  \end{tabular}
\end{center}
\caption{Low-pass approximation of point cloud {\it Bunny}. Plot (a) is the original point cloud with 35,947 points. Plots (b), (c) and (d) show the low-pass approximations with 10, 100 and 400 graph frequency components , respectively. (e) presents the main graph spectral distribution with frequencies higher than 500 omitted as the corresponding magnitudes are around zero.   
}
\label{fig:bandlimited}
\end{figure*}

The smoothed result provides a bandlimited approximation of the original geometric data. 
Fig.~\ref{fig:bandlimited} demonstrates an example of the bandlimited approximation of the 3D coordinates of point cloud {\it Bunny} (35947 points) \cite{levoy2017stanford} with 10, 100 and 400 graph frequencies, respectively.  
Specifically, we construct a $K$-NN graph ($K=10$) on the point cloud and compute the corresponding GFT. 
Then we set the respective bandwidth for low-pass filtering as in \eqref{eq:GFT_filtering_matrix} and \eqref{eq:low_pass_thresholding}.  
One can observe that the first 10 low-frequency components are able to represent the rough shape, with finer details becoming more apparent with additional graph frequencies.     
This validates the assertion that the GFT achieves energy compaction for geometric data.

Another simple choice is a Haar-like low-pass graph filter as discussed in \cite{chen2017fast}, with the graph frequency response as 
\begin{equation}
    \hat{h}(\lambda_k) = 1 - {\lambda_k}/{\lambda_{\mathrm{max}}},
\end{equation}
where $\lambda_{\mathrm{max}}=\lambda_{N}$ is the maximum eigenvalue for normalization. 
As $\lambda_{k-1} \leq \lambda_{k}$, we have $\hat{h}(\lambda_{k-1}) \geq \hat{h}(\lambda_k)$. 
As such, low-frequency components are preserved while high-frequency components are attenuated. 

\subsubsection{High-Pass Graph Spectral Filtering}
In contrast to low-pass filtering, high-pass filtering eliminates low-frequency components and detects large variations in geometric data, such as geometric contours or texture variations. 
A simple design is a Haar-like high-pass graph filter with the following graph frequency response
\begin{equation}
    \hat{h}(\lambda_k) = {\lambda_k}/{\lambda_{\mathrm{max}}}.
\end{equation}
As $\lambda_{k-1} \leq \lambda_{k}$, we have $\hat{h}(\lambda_{k-1}) \leq \hat{h}(\lambda_k)$. 
This indicates that lower-frequency responses are attenuated while high-frequency responses are preserved.  

\subsubsection{Graph Spectral Filtering with a Desired Distribution}
\label{sssec:spectral_filter_desired}
We may also design a desirable spectral distribution and then use graph filter coefficients to fit this distribution. 
For example, an $L$-length graph filter is in the form of a diagonal matrix:
\begin{eqnarray}
\hat{h}(\bfLambda) 
  \ = \ 
  \begin{bmatrix}
    \sum_{k = 0} ^{L-1}  \hat{h}_{k} \lambda_1^{k} &    &   \\
     &  \ddots &   \\
   &  & \sum_{k = 0} ^{L-1} \hat{h}_{k} \lambda_N^{k}
  \end{bmatrix}, 
\end{eqnarray}
where $\bfLambda$ is a diagonal matrix containing eigenvalues of graph Laplacian $\mathbf{L}$ as discussed in Section~\ref{subsec:GFT}, and $\hat{h}_{k}$ is the filter coefficients.  
If the desirable response of the $i$-th graph frequency is $c_i$, we let
\begin{equation}
  \hat{h}(\lambda_i) \ = \ \sum_{k = 0} ^{L-1} \hat{h}_{k} \lambda_i^{k} \
  = \ c_i,
\end{equation}
and solve a set of linear equations to obtain the graph filter
coefficients, $\hat{h}_{k}$. 
An alternative to construct such a graph filter is the Chebyshev polynomial coefficients introduced in~\cite{hammond2011wavelets}.

\vspace{-0.1in}

\subsection{Applications in Geometric Data}

Having discussed graph spectral filtering, we review some representative applications of spectral-domain GSP methods for geometric data, including restoration and compression. 

\subsubsection{Geometric Data Restoration}
Low-pass graph spectral filtering is often designed for geometric data restoration such as denoising. 
As demonstrated in the example of Fig.~\ref{fig:bandlimited}, clean geometric data such as point clouds are dominated by low-frequency components in the GFT domain.
Hence, a carefully designed low-pass filter is able to remove high-frequency components that are likely introduced by noise or outliers.

Based on this principle, Hu \etal proposed depth map denoising by iterative thresholding in the GFT domain \cite{hu2013depth}. 
To jointly exploit local smoothness and non-local self-similarity of a depth map, they cluster self-similar patches and compute an average patch, from which a graph is deduced to describe correlations among adjacent pixels. 
Then self-similar patches are transformed into the same GFT domain, where the GFT basis is computed from the derived correlation graph. 
Finally, iterative thresholding in the GFT domain is performed as the ideal low-pass graph filter in \eqref{eq:low_pass_thresholding} to enforce group sparsity. 

Rosman \etal proposed spectral point cloud denoising based on the non-local framework as well \cite{rosman2013patch}. 
Similar to Block-Matching 3D filtering (BM3D) \cite{dabov2006image}, they group similar surface patches into a collaborative patch and compute the graph Laplacian from this grouping. 
Then they perform shrinkage in the GFT domain by a low-pass filter similar to \eqref{eq:low_pass_thresholding}, which leads to denoising of the collaborative patch. 

In contrast, high-pass graph filtering can be used to detect contours in 3D point cloud data as these are usually represented by high-frequency components.
For instance, Chen \etal proposed a high-pass graph-filtering-based resampling strategy to highlight contours for large-scale point cloud visualization; the same technique can also be used to extract key points for accurate 3D registration \cite{chen2017fast}.
 
\subsubsection{Geometric Data Compression}
\label{subsubsec:compression}
Transform-based coding is generally a low-pass filtering approach. 
When coding piece-wise smooth geometric data, the GFT produces small or zero high-frequency components since it does not filter across boundaries, thus leading to a compact representation in the transform domain. 
Further, as discussed in Section~\ref{ssec:con_inter}, the GFT approximates the KLT in terms of optimal signal decorrelation under a family of statistical processes. 

Graph transform coding is suitable for depth maps due to the piece-wise smoothness. 
Shen \etal first introduced a graph-based representation for depth maps that is adaptive to depth discontinuities, transforming the depth map into the GFT domain for compression and outperforming traditional DCT coding \cite{shen2010edge}. 
Variants of this work include \cite{cheung2011depth,kim2012graph}. 
To further exploit the piece-wise smoothness of depth maps, Hu \etal proposed a multi-resolution compression framework, where boundaries are encoded in the original high resolution to preserve sharpness, and smooth surfaces are encoded at low resolution for greater efficiency \cite{hu2012depth,hu2014multiresolution}. 
It is also shown in \cite{hu2014multiresolution} that the GFT approximates the KLT under a model specifically designed to characterize piece-wise smooth signals.
% thereby minimizing the total signal representation cost of each pixel block.
Other graph transforms for depth map coding include Generalized Graph Fourier Transforms (GGFTs) \cite{hu2015intra} and lifting transforms on graphs \cite{chao2015edge}. 

3D point clouds also exhibit certain piece-wise smoothness in both geometry and attributes. 
Zhang \etal first proposed using graph transforms for attribute compression of static point clouds~\cite{zhang2014point}, where graphs are constructed over local neighborhoods in the point cloud by connecting nearby points, and the attributes are treated as graph signals. 
The graph transform decorrelates the signal and was found to be much more efficient than traditional octree-based coding methods. 
Other follow up work includes graph transforms for sparse point clouds \cite{cohen2016attribute,cohen2016point}, graph transforms with optimized Laplacian sparsity \cite{shao2017attribute}, normal-weighted graph transforms \cite{xu2018cluster}, Gaussian Process Transform (GPT) \cite{de2017transform}, and graph transforms for the enhancement layer \cite{de2018graph}.

In 4D dynamic point clouds, motion estimation becomes necessary to remove the temporal redundancy \cite{thanou2016graph,anis2016compression,xu2019predictive}. 
Thanou \etal represented the time-varying geometry of dynamic point clouds with a set of graphs,
and considered 3D positions and color attributes of the point clouds as signals on the vertices of the graphs \cite{thanou2016graph}. 
Motion estimation is then cast as a feature matching problem between successive graphs based on spectral graph wavelets.   
Dynamic point cloud compression remains a challenging task as each frame is irregularly sampled without any explicit temporal pointwise correspondence with neighboring frames.

\vspace{-0.05in}
\section{Nodal-Domain GSP Methods for \\ Geometric Data}
\label{sec:nodal}
\begin{table*}[htbp]
  \centering\scriptsize
  \caption{Properties of different Graph Smoothness Regularizers (GSR).}
    \begin{tabular}{c|c|c|c|c}
    \hline
    \textbf{Graph Smoothness Regularizer (GSR)} & \textbf{Math Expression} & \textbf{Dynamic} & \textbf{Typical Solver} & \textbf{Typical Works} \bigstrut\\
    \hline
    \hline
    Graph Laplacian Regularizer (GLR) & $\sum_{i \sim j} a_{i,j} \cdot (x_i - x_j)^2$     & No    & Direct Solver / CG to \eqref{eq:closed_form_GLR} & \cite{zeng20183d,pang2017,hu2019feature} \bigstrut\\
    \hline
    Reweighted Graph Laplacian Regularizer (RGLR) & $\sum_{i \sim j} a_{i,j}(x_i, x_j) \cdot (x_i - x_j)^2$     & Yes   & Proximal Gradient & \cite{dinesh2020point} \bigstrut\\
    \hline
    Graph Total Variation (GTV) & $\sum_{i \sim j} a_{i,j} \cdot |x_i - x_j|$     & No    & Primal-Dual Method & \cite{Elmoataz2008TIP,Schoenenberger2015Graph} \bigstrut\\
    \hline
    Reweighted Graph Total Variation (RGTV) & $\sum_{i \sim j} a_{i,j}(x_i,x_j) \cdot |x_i - x_j|$     & Yes   & ADMM  & \cite{bai2018graph} \bigstrut\\
    \hline
    \end{tabular}%
  \label{tab:GSR}
\end{table*}%

\vspace{-0.1in}
\subsection{Basic Principles}
\label{subsec:principle_nodal}
In contrary to spectral-domain GSP methods, this class of methods performs filtering on geometric data locally in the nodal domain, which is often computationally efficient and thus, amenable to \textit{large-scale} data. 

Let $\cN_{n, p}$ be a set of $p$-hop neighborhood nodes of the $n$-th vertex, whose cardinality often varies according to $n$. 
Nodal-domain filtering is typically defined as a linear combination of local neighboring vertices
\begin{equation}
\label{eq:nodal_filter_basic}
y_n := \sum_{j \in \cN_{n, p}} h_{n,j} x_j,
\end{equation}
where $h_{n,j}$ denotes filter coefficients of the graph filter. 
Since $\cN_{n, p}$ is node-dependent, $h_{n,j}$ needs to be properly defined according to $n$. 

Typically, $h_{n,j}$ may be parameterized as a function of the adjacency matrix $\A$:
\begin{equation}
\label{eq:nodal_filter_A}
    \y = h(\A) \x, 
\end{equation}
where
\begin{equation}
  h(\A) \ = \ \sum_{k = 0} ^{K-1}  h_{k} \A^{k} = h_0 \I + h_1 \A + \ldots + h_{K-1} \A^{K-1}. 
  \label{eq:nodal_filtering_A}
\end{equation}
Here $h_{k}$ is the $k$-th filter coefficient that quantifies the contribution from the $k$-hop neighbors, and $K$ is the length of the graph filter. 
$\A^k$ determines the $k$-hop neighborhood by definition, thus a higher-order corresponds to a larger filtering range in the graph vertex domain. 
When operating $\A$ on a graph signal, it computes the average of the neighboring signal of each vertex, which is essentially a low-pass filter. 

$\A$ can be replaced by other graph operators such as the graph Laplacian $\mathbf{L}$:
\begin{equation}
  h(\mathbf{L}) \ = \ \sum_{k = 0} ^{K-1}  h_{k} \mathbf{L}^{k} = h_0 \I + h_1 \mathbf{L} + \ldots + h_{K-1} \mathbf{L}^{K-1}. 
  \label{eq:nodal_filtering_L}
\end{equation}
When operating $\mathbf{L}$ on a graph signal, it sums up the signal difference between each vertex and its neighbors, which is essentially a high-pass filter. 

\vspace{-0.1in}
\subsection{Nodal-domain Optimization}
\label{subsec:nodal_optimization}

Besides direct filtering as in \eqref{eq:nodal_filter_basic} or \eqref{eq:nodal_filter_A}, nodal-domain filtering often employs graph priors for regularization.  
Graph Smoothness Regularizers (GSRs), which introduce prior knowledge about smoothness in the underlying graph signal, play a critical role in a wide range of inverse problems, such as depth map denoising \cite{pang2017,hu2013depth}, point cloud denoising \cite{zeng20183d,hu2019feature}, and inpainting \cite{hu2019local}. 

\subsubsection{Formulation}
In general, the formulation to restore a geometric datum ${\bf x}$ with a signal prior, \eg, the GSR, is given by the following maximum a posteriori optimization problem:
\begin{equation}\label{eq:prior_problem}
    {\bf x}^{\star} = \arg\min\limits_{\bf x}\ \left \| {\bf y}-H({\bf x} )\right \|_2^2 + \mu\cdot{\text{GSR}{(\bf x,\mathcal{G})}},
\end{equation}
where $\mathbf{y}$ is the observed signal and $H(\cdot)$ is a degradation operator (\emph{e.g.}, down-sampling) defined over $\bf x$. The first term in \eqref{eq:prior_problem} is a data fidelity term; $\mu \in \mathbb{R}$ balances the importance between the data fidelity term and the signal prior. 

Next, we discuss two classes of commonly used GSRs---Graph Laplacian Regularizer (GLR) and Graph Total Variation (GTV), as well as techniques to solve \eqref{eq:prior_problem} with these priors. 
The property comparison of different GSRs is summarized in Table~\ref{tab:GSR}.

\subsubsection{Graph Laplacian Regularizer (GLR)}
\label{subsubsec:GLR}
The most commonly used GSR is the GLR. 
Given a graph signal $ \x $ residing on the vertices of $ \mathcal{G} $ encoded in the graph Laplacian $\mathbf{L}$, the GLR can be expressed as
\begin{equation}
	\x^{\top} \mathbf{L} \x =\sum_{i \sim j} a_{i,j} \cdot (x_i - x_j)^2,
	\label{eq:prior}
\end{equation}
where $i\sim j$ means vertices $i$ and $j$ are connected, implying the underlying points on the geometry are highly correlated.
$a_{i,j}$ is the corresponding element of the adjacency matrix $\mathbf{A}$. 
The signal $\x$ is smooth with respect to $\mathcal{G}$ if the GLR is small, as connected vertices $ x_i $ and $ x_j $ must be similar for a large edge weight $ a_{i,j} $; for a small $ a_{i,j} $, $x_i$ and $x_j$ can differ significantly. 
This prior also possesses an interpretation in the frequency domain:
\begin{align}\label{eq:glr_freq}
\x^{\top} \mathbf{L} \x =
\sum_{k=1}^N \lambda_k \hat{x}_k^2,               
\end{align}
where $\lambda_k$ is the $k$-th eigenvalue of $\mathbf{L}$, and $\hat{x}_k$ is the the $k$-th GFT coefficient. 
In other words, $\hat{x}_k^2$ is the energy in the $k$-th graph frequency for geometric data $\x$. 
Thus, a small $\x^{\top} \mathbf{L} \x$ means that most of the signal energy is occupied by the low-frequency components.

When we employ the GLR as the prior in \eqref{eq:prior_problem} and assume $H(\cdot)$ is differentiable, \eqref{eq:prior_problem} exhibits a closed-form solution. For simplicity, we assume $H = \I$ (\eg, as in the denoising case), then setting the derivative of \eqref{eq:prior_problem} to zero yields 
\begin{equation}
    \x^{\star} = (\I + \mu \mathbf{L})^{-1} \y,
    \label{eq:closed_form_GLR}
\end{equation}
which is a set of linear equations and can be solved directly or with conjugate gradient (CG) \cite{axelsson1986rate}.
As $\mathbf{L}$ is a high-pass operator, the solution in \eqref{eq:closed_form_GLR} is essentially an adaptive low-pass filtering result from the observation $\y$. This can also be indicated by the corresponding graph spectral response: 
\begin{equation}
    \hat{h}(\lambda_k) = 1/(1 + \mu \lambda_k),
    \label{eq:GLR_spectral_response}
\end{equation}
which is a low-pass filter since smaller $\lambda_k$'s correspond to lower frequencies. 
As described in Section~\ref{sssec:spectral_low_pass}, the low-pass filtering will lead to smoothed geometric data with the underlying shape retained. 

Further, as discussed in Section~\ref{ssec:con_inter}, the graph Laplacian operator converges to the Laplace-Beltrami operator on the geometry in the continuous manifold when the number of samples tends to infinity. 
We can also interpret the GLR from a continuous manifold perspective. 
According to \cite{hein2007graph}, given a Riemannian manifold $\cal M$ (or surface) and a set of $N$ points uniformly sampled on $\cal M$, an $\epsilon$-neighborhood graph $\cal G$ can be constructed with each vertex corresponding to one sample on $\cal M$.
For a function $x$ on manifold $\cal M$ and its discrete samples $\x$ on graph $\cal G$ (a graph signal), under mild conditions,
\begin{equation}\label{eq:glr_conv}
\underset{\substack{N \rightarrow \infty\\ \epsilon \rightarrow 0}} {\lim}   \x^{\top} \mathbf{L} \x 
\sim \frac{1}{|\mathcal{M}|} \int_{\mathcal{M}} \|\nabla_{\mathcal{M}} x(\mathbf{s})\|_2^2 d \mathbf{s},
\end{equation}
where $\nabla_{\cal M}$ is the gradient operator on manifold $\cal M$, and {\bf s} is the natural volume element of $\cal M$ \cite{hein2007graph}.
In other words, the GLR now converges to a smoothness functional defined on the associated Riemannian manifold.
The relationship \eqref{eq:glr_conv} reveals that the GLR essentially regularizes graph signals with respect to the underlying manifold geometry, which justifies the usefulness of the GLR \cite{hein2007manifold}.

% Signal-dependent graph smoothness regularizer
In the aforementioned GLR, the graph Laplacian $\mathbf{L}$ is {\it fixed}, which does not promote reconstruction of the target signal with discontinuities if the corresponding edge weights are not very small. 
It is thus extended to \textit{Reweighted} GLR (RGLR) in \cite{liu2016random,pang2017,bai2018graph} by considering $\mathbf{L}$ as a learnable function of the graph signal $\x$. 
The RGLR is defined as    
\begin{equation}
    \x^{\top} \mathbf{L}(\x) \x = \sum\limits_{i \sim j} a_{i,j}(x_i, x_j) \cdot (x_i - x_j)^2,
    \label{eq:rgssp}
\end{equation}
where $a_{i,j}(x_i, x_j)$ can be learned from the data. 
Now we have two optimization variables $\x$ and $a_{i,j}$, which can be optimized alternately via proximal gradient \cite{Parikh31}.  

It has been shown in \cite{bai2018graph} that minimizing the RGLR iteratively can promote piece-wise smoothness in the reconstructed graph signal $\x$, assuming that the edge weights are appropriately initialized. 
Since geometric data often exhibits piece-wise smoothness as discussed in Section~\ref{subsec:characteristics}, the RGLR helps to promote this property in the reconstruction process.

\subsubsection{Graph Total Variation (GTV)}
Another popular line of GSRs generalizes the well-known Total Variation (TV) regularizer \cite{TV} to graph signals, leading to the Graph Total Variation (GTV) and its variants. The GTV is defined as \cite{Couprie2013SIAM}:
\begin{equation}\label{eq:GTV}
    \|{\bf x}\|_{\text{GTV}} = \sum\limits_{i \sim j} a_{i,j} \cdot |x_i - x_j|.
\end{equation}
where $a_{i,j}$ is fixed during the optimization. 
Since the GTV is non-differentiable, \eqref{eq:GTV} has no closed-form solution, but can be solved via existing optimization methods such as the primal-dual algorithm \cite{zhang2011unified}. 

Instead of using fixed $\mathbf{A}$, Bai \textit{et al}. extended the conventional GTV to the Reweighted GTV (RGTV) \cite{bai2018graph}, where graph weights are dependent on $\mathbf{x}$:
\begin{equation}\label{eq:RGTV}
    \|{\bf x}\|_{\text{RGTV}} = \sum\limits_{i \sim j} a_{i,j}(x_i,x_j) \cdot |x_i - x_j|.
\end{equation}
This can be solved by ADMM \cite{wahlberg2012admm} or the algorithm proposed in \cite{bai2018graph}. 

The work of \cite{bai2018graph} also provides spectral interpretations of the GTV and RGTV by rewriting them as $\ell_1$-Laplacian operators on a graph. The spectral analysis demonstrates that the GTV is a stronger PWS-preserving filter than the GLR, and the RGTV has desirable properties including robustness to noise and blur and promotes sharpness. 
Hence, the RGTV is advantageous to boosting the piece-wise smoothness of geometric data. 

\vspace{-0.1in}
\subsection{Applications in Geometric Data}
\begin{figure*}[t]
  \begin{center}
    \begin{tabular}{ccccc}
    \includegraphics[width=0.17\textwidth]{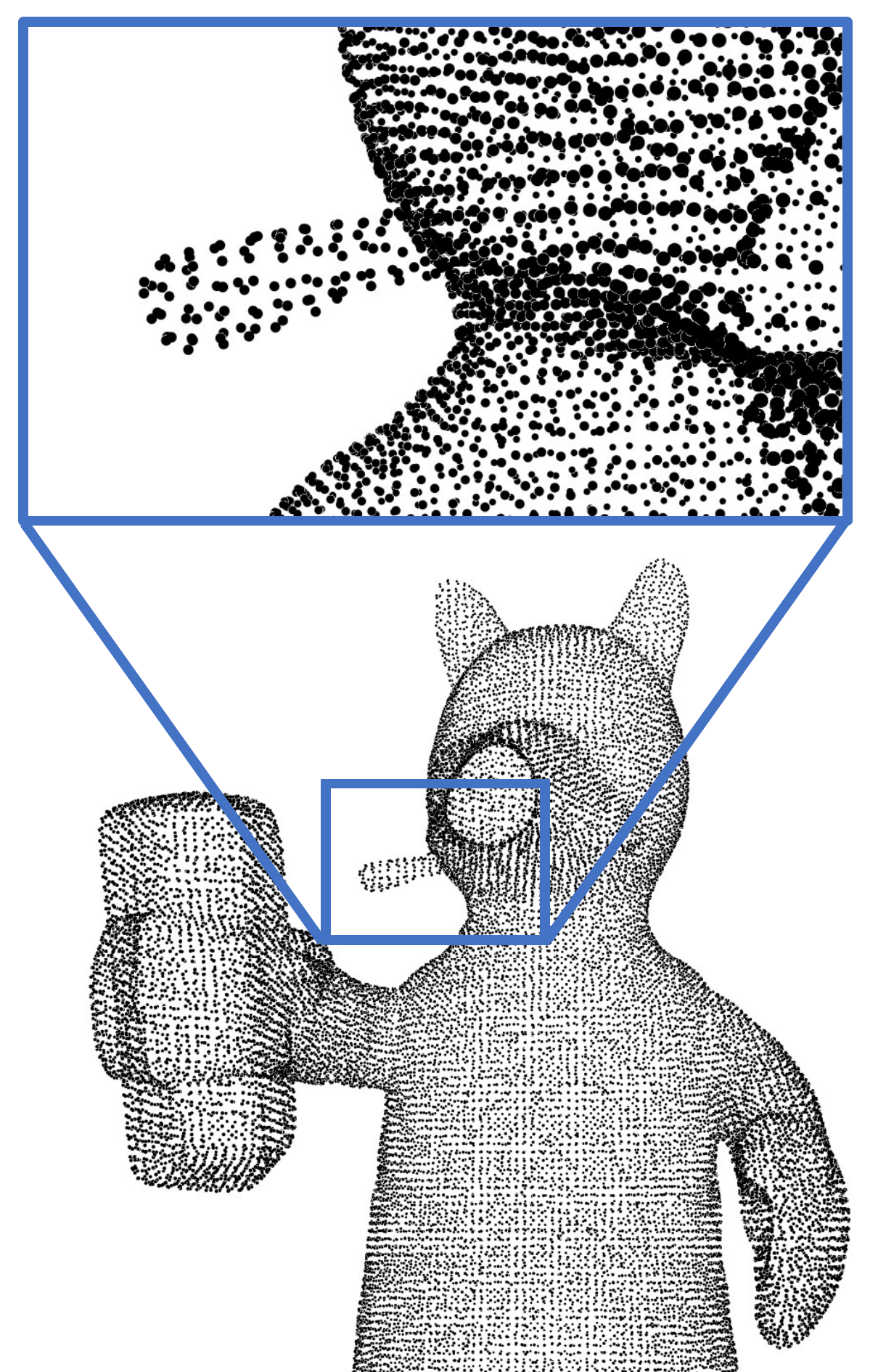} & \includegraphics[width=0.17\textwidth]{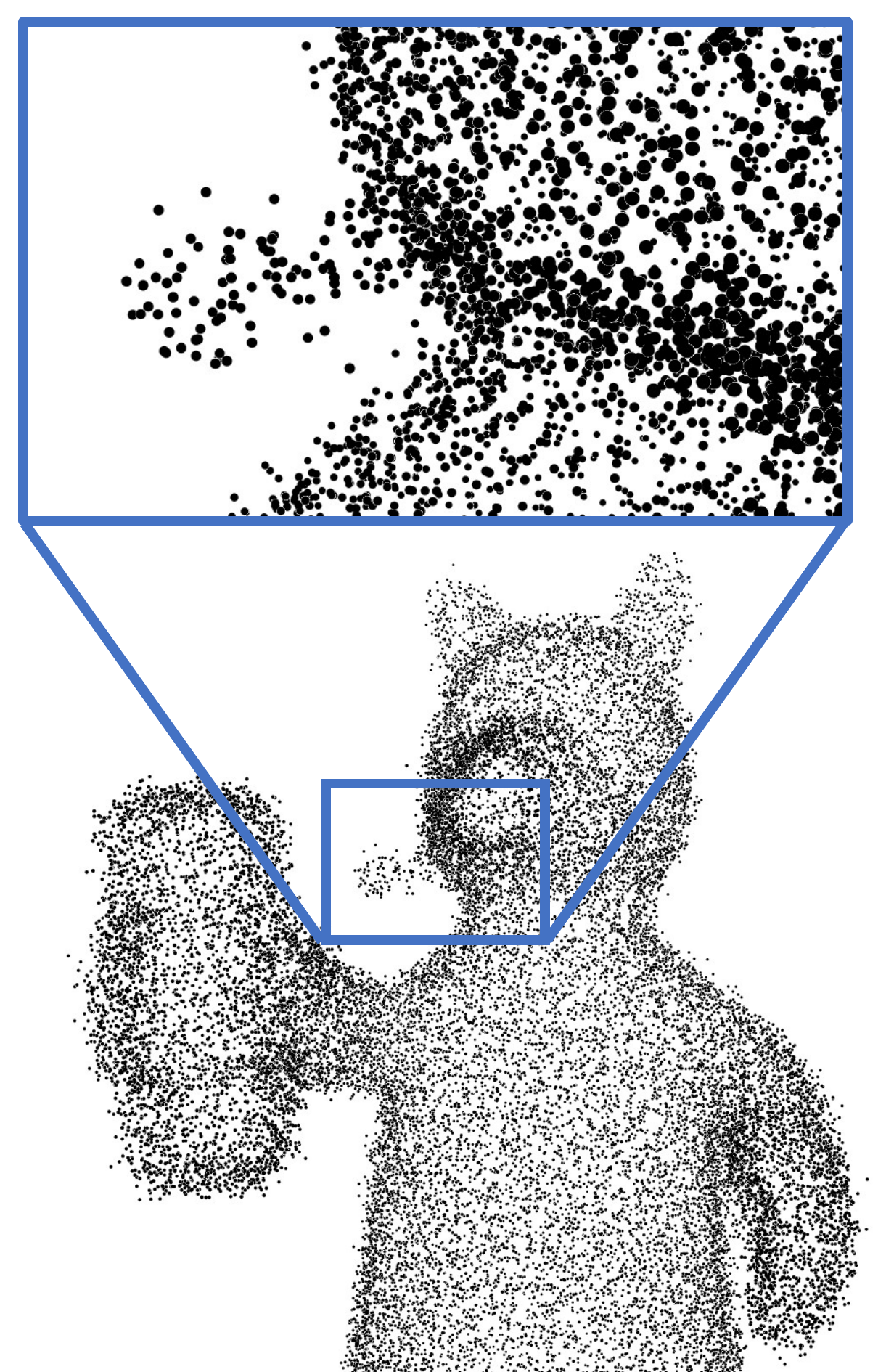} & \includegraphics[width=0.17\textwidth]{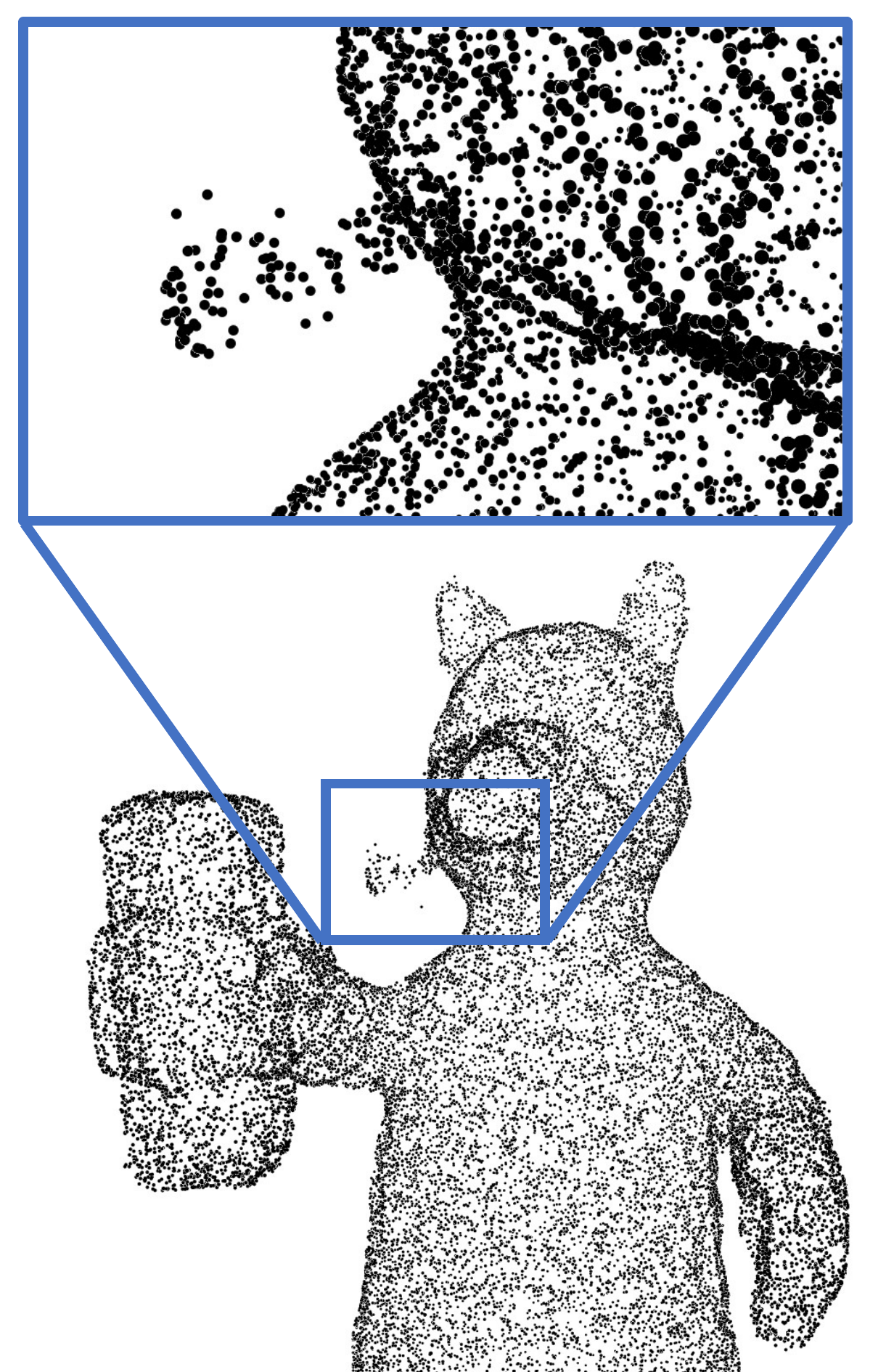} & 
    \includegraphics[width=0.17\textwidth]{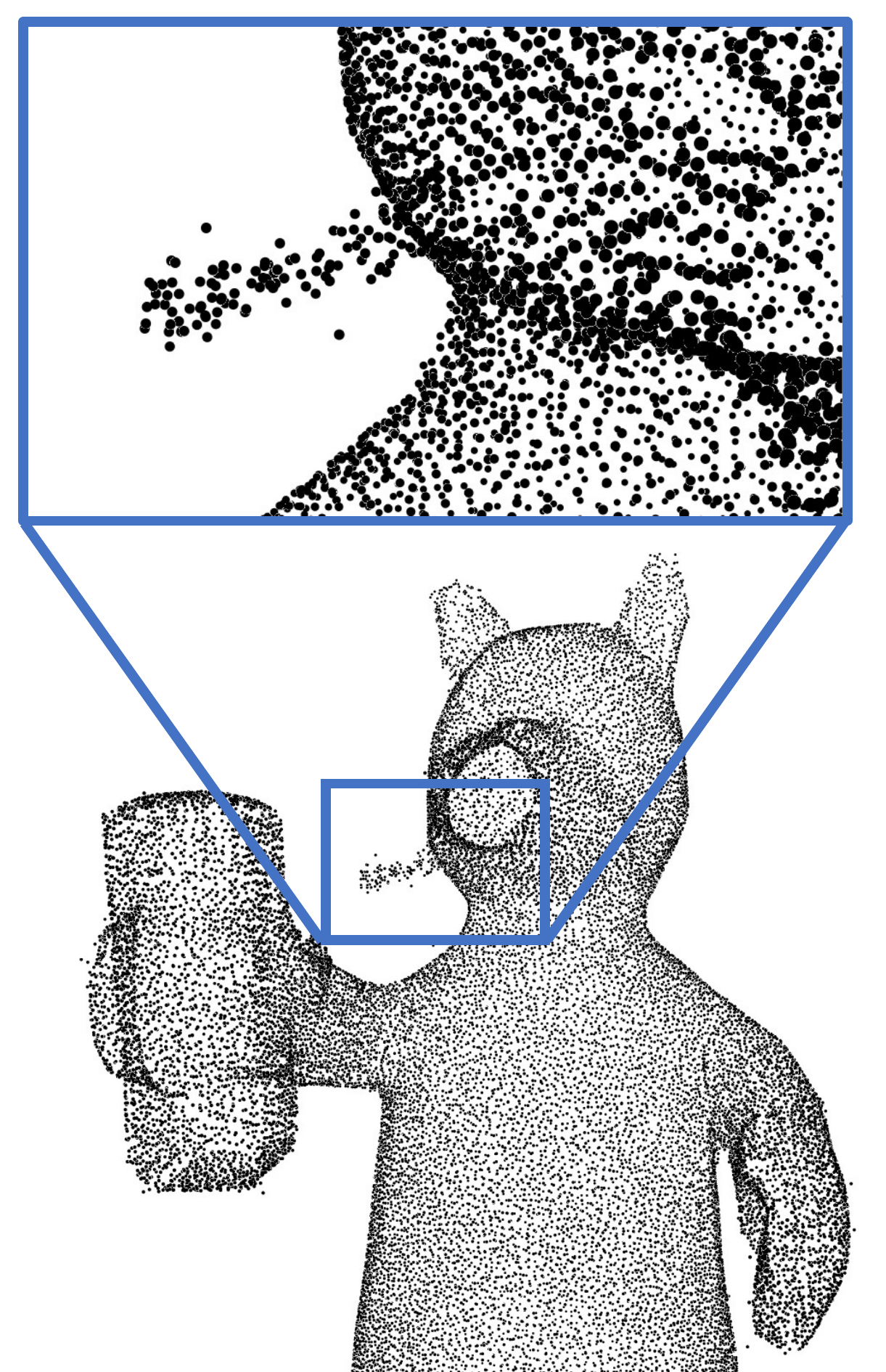} & 
    \includegraphics[width=0.17\textwidth]{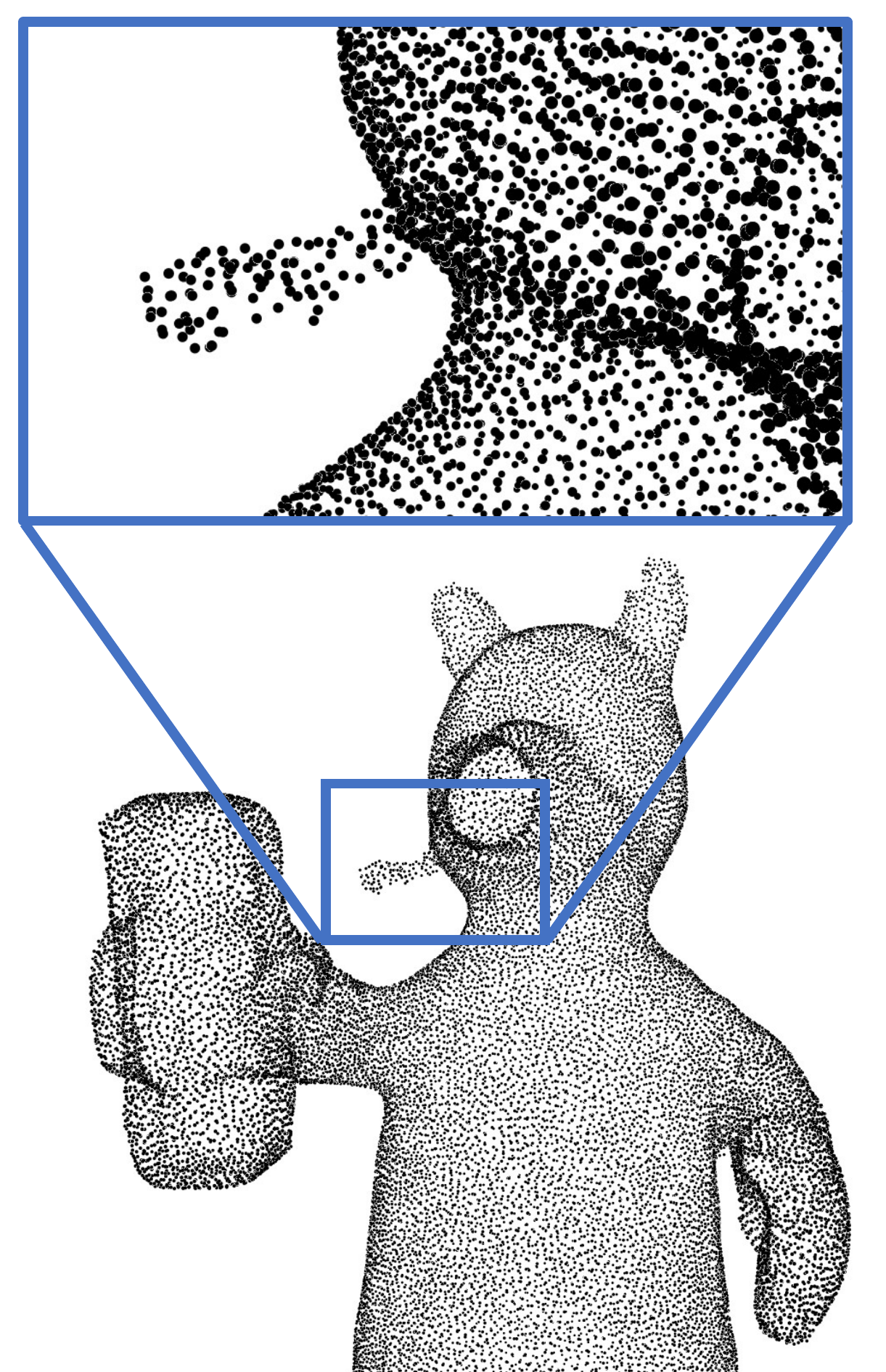}
    \\
    {\small (a) Ground truth.} & {\small (b) Noisy.} &  {\small (c) Spectral-LP. } & {\small (d) Nodal-\cite{zeng20183d}.} & {\small (e) Nodal-\cite{hu2019feature}.}  
  \end{tabular}
\end{center}
\caption{Point cloud denoising results with Gaussian noise $\sigma=0.04$ for \textit{Quasimoto} \cite{berger2013benchmark}: (a) The ground truth; (b) The noisy point cloud; (c) The denoised result by graph spectral low-pass (LP) filtering that we implement according to \eqref{eq:low_pass_thresholding}; (d) The denoised result by a nodal-domain GSP method in \cite{zeng20183d}; (e) The denoised result by a nodal-domain GSP method in \cite{hu2019feature}.}
\label{fig:denoising_results}
\end{figure*}

In the following, we review a few works on geometric data restoration with nodal-domain GSP methods.
First, we present a few applications recovering geometric data with the simple-yet-effective GLR, and then extend our scope to more advanced graph smoothness regularizers.
\subsubsection{Geometric Data Restoration with the GLR}
To cope with various geometric data restoration problems, GLR-based methods place more emphasis on the choice of the neighborhood graph and the algorithm design.
For instance, to remove additive white Gaussian noise (AWGN) from depth images, Pang and Cheung~\cite{pang2017} adopted the formulation in \eqref{eq:prior_problem} with the GLR.
To understand the behavior of the GLR for 2D depth images, \cite{pang2017} performs an analysis in the continuous domain, leading to an $\epsilon$-neighborhood graph (Section~\ref{ssec:graph_rep}) which not only \emph{smooths} out noise but also \emph{sharpens} edges.

Zeng~{\it et~al.} \cite{zeng20183d} applied the GLR for point cloud denoising.
In contrast to \cite{pang2017}, they first formulated the denoising problem with a low dimensional manifold model (LDMM) \cite{osher2017low}.
The LDMM prior suggests that the clean point cloud patches are samples from a low-dimensional manifold embedded in the high dimensional space, though it is non-trivial to minimize the dimension of a Riemannian manifold.
With \eqref{eq:glr_conv} and tools from differential geometry \cite{helgason1979differential}, it is possible to ``convert'' the LDMM signal prior to the GLR.
Hence, the problem of minimizing the manifold dimension is approximated by iteratively solving a quadratic program with the GLR.

Instead of constructing the underlying graph with pre-defined edge weights from hand-crafted parameters, Hu \etal proposed feature graph learning by minimizing the GLR using the Mahalanobis distance metric matrix $\M$ as a variable, assuming a feature vector per node is available \cite{hu2019feature}. 
Then the graph Laplacian $\mathbf{L}$ becomes a function of $\M$, {\it i.e.}, $\mathbf{L}(\M)$.   
A fast algorithm with the GLR is presented and applied to point cloud denoising, where the graph for each set of self-similar patches is computed from 3D coordinates and surface normals as features.

\subsubsection{Geometric Data Restoration with Other GSRs}
Despite the simplicity of the GLR, applying it for geometric data restoration involves sophisticated graph construction or algorithmic procedures.
This has motivated the development of other geometric data restoration methods using various GSRs that are tailored to specific restoration tasks.

To remove noise on point clouds, the method proposed in~{\it et~al.}~\cite{Schoenenberger2015Graph} first assumes smoothness in the gradient $\nabla_{\mathcal G} {\bf Y}$ of the point cloud ${\bf Y}$ on a graph $\mathcal G$, leading to a Tikhonov regularization $\text{GSR}_{\text{Tik}}(\Y)=\|\nabla_{\mathcal G} \Y\|_2^2$ which is equivalent to the simple GLR. 
The method further assumes the underlying manifold of the point cloud to be \emph{piece-wise smooth} rather than smooth, and then replaces the Tikhonov regularization with the GTV regularization \eqref{eq:GTV}, {\it i.e.}, $\text{GSR}_{\text{TV}}(\Y)=\|\nabla_{\mathcal G} \Y\|_1$.
In \cite{Elmoataz2008TIP}, Elmoataz \etal also applied the GTV for mesh filtering to simplify 3D geometry. 

In \cite{dinesh2020point}, Dinesh \etal applied the RGTV \eqref{eq:RGTV} to regularize the surface normal for point cloud denoising, where the edge weight between two nodes is a function of the normals.  
Moreover, they established a linear relationship between normals and 3D point coordinates via bipartite graph approximation for ease of optimization.
To perform point cloud inpainting, Hu~{\it et~al.}~\cite{hu2019local} also applied a modified GTV called Anisotropic GTV (AGTV) as a metric to measure the similarity of point cloud patches.

\subsubsection{Restoration with Nodal-domain Filtering}
Solving optimization problems can be formidable and sometimes even impractical.
An alternative strategy for geometric data recovery is to perform \emph{filtering} in the nodal-domain as discussed in Section~\ref{subsec:principle_nodal}.
Examples include point cloud re-sampling \cite{chen2017fast} and depth image enhancement \cite{wang2014graph}.
Essentially, nodal-domain filtering aims at ``averaging'' the samples of a graph signal adaptively, either locally or non-locally.

\vspace{-0.1in}
\subsection{Discussion on Spectral- and Nodal-Domain GSP Methods}

There exists a close connection between spectral-domain methods and nodal-domain ones, and as discussed earlier, there is a correspondence between spatial graph filters and their graph spectral response.
As an example, consider the filter in \eqref{eq:nodal_filtering_L}, where $\mathbf{L}^k = (\U \bfLambda \U^{\top})^k = \U \bfLambda^k \U^{\top}$ since $\U^{\top} \U = \I$. It follows that \eqref{eq:nodal_filtering_L} can be rewritten as  
\begin{equation}
    h(\mathbf{L}) = \U h(\bfLambda) \U^{\top},
\end{equation}
which corresponds to a graph spectral filter with $h(\bfLambda)$ as the spectral response in \eqref{eq:GFT_filtering_matrix}. 
Another example is the spectral response of the solution to a GLR-regularized optimization problem, as presented in \eqref{eq:GLR_spectral_response}. 

In addition, some nodal-domain graph filtering methods are approximations of spectral-domain filtering, including polynomial approximations of graph spectral filters like Chebyshev polynomials \cite{hammond2011wavelets,gadde2013bilateral,tian2014chebyshev} for depth map enhancement, as well as lifting transforms on graphs \cite{chao2015edge} for depth map coding. 

Comparing the two methods, as mentioned earlier, spectral methods entail higher computational complexity due to eigen-decomposition, whereas spatial methods avoid such complexity and are thus more amenable to large-scale geometric data. 
Also, graph transforms employed in spectral methods are mostly {\it global} transforms, which capture the global features. 
In contrast, spatial methods are often applied to capture {\it local} features in graph signals. 
Taking point cloud denoising as an example, we show  denoising results in Fig.~\ref{fig:denoising_results} comparing one graph spectral low-pass filtering method that we implement according to \eqref{eq:low_pass_thresholding} as well as two state-of-the-art nodal-domain GSP methods \cite{zeng20183d,hu2019feature}. 
We can observe from these results that the nodal-domain methods reconstruct local structural features better, including fine components such as the tobacco pipe.

\vspace{-0.05in}
\section{GSP-based Interpretation for Graph Neural Networks}
\label{sec:gnn}
\begin{figure}[t]
  \begin{center}
    \begin{tabular}{c}
    \includegraphics[width=0.35\textwidth]{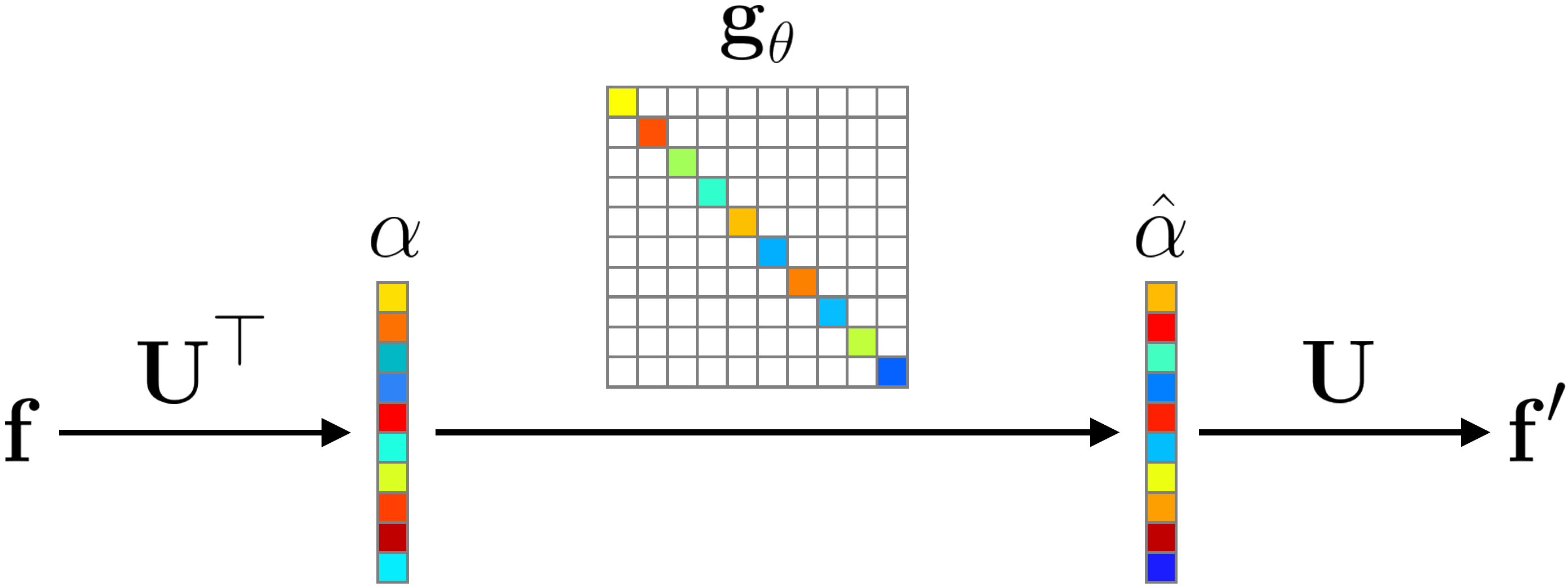}\\
    {\small (a)~Spectral graph convolution.}\\
    \\
    \includegraphics[width=0.2\textwidth]{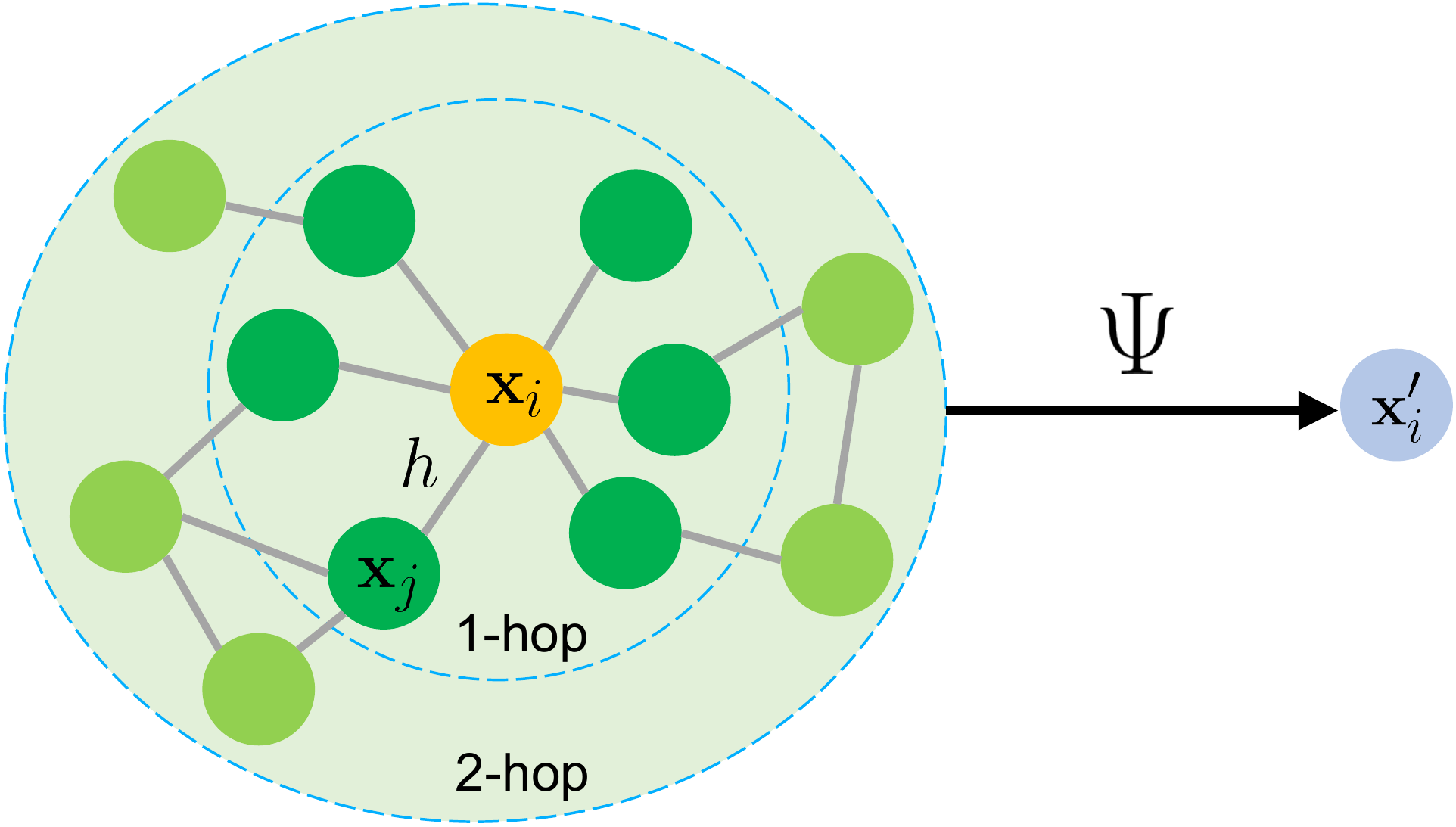}\\
    {\small (b)~Spatial graph convolution.}
  \end{tabular}
\end{center}
\vspace{-0.1in}
\caption{Graph convolution operations.}
\label{fig:gnn_filt}
\end{figure}

The aforementioned GSP methods are model-based and built upon prior knowledge and characteristics of geometric data, which usually perform robustly, \eg, a depth map denoising algorithm would still perform reasonably on natural images.  
However, model-based approaches lack flexibility as they are built upon prior observations, \eg, the GSRs in Section~\ref{subsec:nodal_optimization}~\cite{zeng2019deep}.
In contrast, learning-based methods infer filter parameters in a data-driven manner, which are highly flexible, such as the recently developed geometric deep learning \cite{bronstein2017geometric}.
On the other hand, compared to the hand-crafted assumptions made in model-based approaches, learning-based approaches effectively learn to abstract high-level (or semantic) features with a training process \cite{lecun2015deep}.
Consequently, they are more suitable for high-level applications such as segmentation~\cite{qi20173d} and classification~\cite{zhang2018graph}.

Convolutional Neural Networks (CNNs) have shown to be extremely effective for a wide range of imaging tasks but have been designed to process data defined on regular grids, where spatial relationships between data samples (\eg, top, bottom, left and right) are uniquely defined. In order to leverage such networks for geometric data, some prior works transform irregular geometric data to regular 3D voxel grids or collections of 2D images before feeding them to a neural network \cite{su2015multi, ma2018learning} or impose certain symmetries in the network computation (\eg, PointNet \cite{qi2017pointnet}, PointNet++ \cite{qi2017pointnet++}, PointCNN \cite{li2018pointcnn}). As discussed in Section~\ref{subsec:non_graph}, non-graph representations are sometimes redundant, inaccurate or deficient in data structural description.

In contrast, GSP provides efficient filtering and sampling of such data with insightful spectral interpretation, which is able to generalize the key operations ({\it e.g.}, convolution and pooling) in a neural network to irregular geometric data. 
For example, graph convolution can be defined as graph filtering either in the spectral or the spatial domain.    
This leads to the recently developed Graph Neural Networks (GNNs) (see \cite{bronstein2017geometric} and references therein), which generalize CNNs to unstructured data.
GNNs have achieved success in both analysis and synthesis of geometric data.
The input geometric features at each point (vertex) of GNNs are usually assigned with coordinates, laser intensities or colors, while features at each edge are usually assigned with geometric similarities between two connected points. 

Nonetheless, learning-based methods are facing common issues such as {\it interpretability}, {\it robustness} and {\it generalization} \cite{zeng2019deep,valsesia2020deep}.
In the remainder of this section, we will particularly discuss the interpretability of GNNs from the perspective of GSP for geometric data, which is expected to inspire more interpretable, robust, and generalizable designs of GNNs.   

\vspace{-0.1in}
\subsection{Interpreting Graph Convolution with GSP}

GSP tools, particularly graph filters, inspire some early designs of basic operations in GNNs, including spectral graph convolution and spectrum-free graph convolution. 
In addition, GSP provides interpretation for spatial graph convolution from the perspective of spatial graph filtering. 

\subsubsection{Spectral Graph Convolution} 
As there is no clear definition of shift-invariance over graphs in the nodal domain, one may define graph convolution in the spectral domain via graph transforms according to the Convolution Theorem. 
That is, the graph convolution of signal $\f \in \mathbb{R}^{N}$ and filter $\g \in \mathbb{R}^{N}$ in the spectral domain with respect to the underlying graph $\mathcal{G}$ can be expressed as the element-wise product of their graph transforms:
\begin{equation}
    \g \star_{\mathcal{G}} \f = \U (\U^{\top} \g \odot \U^{\top} \f), 
\end{equation}
where $\U$ is the GFT basis and $\odot$ denotes the Hadamard product. Let $\g_\theta = \text{diag}(\U^{\top} \g)$, the graph convolution can be simplified as
\begin{equation}
    \g \star_{\mathcal{G}} \f = \U \g_\theta \U^{\top} \f.
    \label{eq:spectral_conv}
\end{equation}
The key difference in various spectral GNN methods is the choice of filter $\g_\theta$ which captures the holistic appearance of the geometry. 
In an early work \cite{bruna2013spectral}, $\g_\theta = \Theta$ is a learnable diagonal matrix.

As schematically shown in Fig.~\ref{fig:gnn_filt}(a), the spectral-domain graph convolution \eqref{eq:spectral_conv} is essentially the {\it spectral graph filtering} defined in \eqref{eq:GFT_filtering_matrix} if the diagonal entries of $\g_\theta$ are the graph frequency response $\hat{h}(\lambda_k)$. 
As $\g_\theta$ is often learned so as to adapt to various tasks, it is analogous to the graph spectral filtering with desired distribution as discussed in Section~\ref{sssec:spectral_filter_desired}.  
Hence, we are able to interpret spectral-domain graph convolution via spectral graph filtering for geometric data processing. 

\subsubsection{Spectrum-free Graph Convolution} 
It has been noted earlier that the eigen-decomposition required by spectral-domain graph convolution incurs relatively high computational complexity. 
However, one may parameterize the filter using a smooth spectral transfer function $\Theta(\bfLambda)$ \cite{henaff2015deep}. 
One choice is to represent $\Theta(\bfLambda)$ as a $K$-degree polynomial, such as the Chebyshev polynomial which approximates the graph kernel well \cite{hammond2011wavelets}:  
\begin{equation}
    \Theta(\bfLambda) = \sum_{k=0}^{K-1} T_k(\tilde{\bfLambda}),
    \label{eq:cheby_spectral}
\end{equation}
where $T(\cdot)$ denotes the Chebyshev polynomial. 
It is defined as $T_0(\tilde{\bfLambda}) = \I$, $T_1(\tilde{\bfLambda}) = \tilde{\bfLambda}$, $T_k(\tilde{\bfLambda}) = 2 \tilde{\bfLambda} T_{k-1}(\tilde{\bfLambda}) - T_{k-2}(\tilde{\bfLambda})$. 
$\tilde{\bfLambda}$ denotes the normalized eigenvalues in $[-1,1]$ due to the domain defined by the Chebyshev polynomial.

Combining \eqref{eq:spectral_conv} and \eqref{eq:cheby_spectral}, we have 
\begin{align}
    \g \star_{\mathcal{G}} \f & = \U \Theta(\bfLambda) \U^{\top} \f \\
    & \approx \U \sum_{k=0}^{K} T_k(\tilde{\bfLambda}) \U^{\top} \f = \sum_{k=0}^{K} T_k(\widetilde{\mathbf L}) \f,
    \label{eq:chebnet}
\end{align}
where $\widetilde{\mathbf L} = \U \tilde{\bfLambda} \U^{\top}$ is a normalized graph Laplacian. 
This leads to well-known ChebNet \cite{defferrard2016convolutional}. 

If we only consider $1$-degree Chebyshev polynomial, namely, $K=1$, it leads to the widely used Graph Convolutional Network (GCN) \cite{kipf2016semi}. 
With a series of simplifications and renormalization, the convolutional layer of the GCN takes the form:
\begin{equation}
    \g \star_{\mathcal{G}} \f = \widetilde{\mathbf D}^{-\frac{1}{2}} \widetilde{\mathbf A}\widetilde{\mathbf D}^{-\frac{1}{2}} \Phi,
    \label{eq:gcn}
\end{equation}
where $\widetilde{\mathbf A}=\mathbf A + \mathbf I$ is the renormalized adjacency matrix, and $\widetilde{\mathbf D}$ is the corresponding degree matrix. 
$\Phi$ is a matrix of filter parameters.

While inspired from a graph spectral viewpoint, both the ChebNet and GCN can be implemented in the spatial domain directly, which are thus referred to as \emph{spectrum-free}. 
The spectrum-free convolution in \eqref{eq:chebnet} and \eqref{eq:gcn} is essentially nodal-domain graph filtering presented in \eqref{eq:nodal_filtering_L} and \eqref{eq:nodal_filtering_A}, respectively. 
For instance, the graph convolution in the GCN is a simple one-hop neighborhood averaging.

\subsubsection{Spatial Graph Convolution}\label{subsubsec:spatial_geometric_learning}
Analogous to the convolution in CNNs, spatial graph convolution aggregates the information of neighboring vertices to capture the local geometric structure in the spatial domain, leading to feature propagation over adjacent vertices that enforce the smoothness of geometric data to some extent \cite{simonovsky2017dynamic,monti2017geometric,wang2019dgcnn}. 
Such graph convolution filters over the neighborhood of each vertex in the spatial domain are essentially nodal-domain graph filters from the perspective of GSP.  

As a representative spatial method on point clouds, Wang \etal introduced the concept of {\it edge convolution} \cite{wang2019dgcnn}, which generates edge features that characterize the relationships between each point and its neighbors. 
The edge convolution exploits local geometric structure and can be stacked to learn global geometric properties. 
Let $\x_i \in \RR^d$ and $\x_j \in \RR^d$ denote the graph signal on the $i$-th and $j$-the vertex respectively, the output of edge convolution is:
\begin{equation}\label{eq:gnn_spat}
\x_i' = \Psi_{(i,j) \in \mathcal E} h (\x_i, \x_j) \in \RR^d,
\end{equation}
where $\mathcal E$ is the set of edges and $h(\cdot, \cdot)$ is a generic edge feature function, implemented by a certain neural network. 
$\Psi$ is a generic aggregation function, which could be the summation or maximum operation. 
The operation of \eqref{eq:gnn_spat} is demonstrated in Fig.~\ref{fig:gnn_filt}(b), where we could consider not only the 1-hop neighbors but also 2-hop neighbors or more.

The edge convolution is also similar to the nodal-domain graph filtering: both aggregate neighboring information; further, the edge convolution specifically models each pairwise relationship by a non-parametric function. 

\vspace{-0.1in}
\subsection{Understanding Representation Learning of GNNs with GSP}
GSP tools also provide interpretation for representation learning of GNNs, as discussed below.

\subsubsection{Low-pass Graph Filtering of Features}
Wu~\etal~\cite{wu2019simplifying} propose to simplify GCNs by successively removing nonlinearities in GCNs and collapsing weight matrices between consecutive layers and analyze that this simple graph convolution (SGC) corresponds to a fixed low-pass filter followed by a linear classifier. 
That is, SGC acts as a low-pass filter that produces smooth features over adjacent nodes in the graph. 
As a result, nearby nodes tend to share similar representations and consequently predictions.

Fu \etal \cite{fu2020understanding} show that several popular GNNs can be interpreted as implicitly implementing denoising and/or smoothing of graph signals.
In particular, spectral graph convolutions \cite{defferrard2016convolutional,kipf2016semi} work as denoising node features, while graph attentions \cite{velivckovic2017graph, li2018adaptive}
work as denoising edge weights. 

\subsubsection{Introducing Domain Knowledge via GSP-based Regularization}
\label{subsubsec:domain_knowledge}
Some works provide domain knowledge via GSP-based regularization (\eg, GSRs) for better understanding the representational properties of GNNs. 
For instance, Te \etal proposed a Regularized Graph Convolutional Neural Network (RGCNN) \cite{te2018rgcnn} as one of the first approaches to utilize GNNs for point cloud segmentation, which regularizes each layer by the GLR introduced in Section~\ref{subsubsec:GLR}. 
This prior essentially enforces the features of vertices within each connected component of the graph similar, which is incorporated into the loss function and enables explainable and robust segmentation.
Also, the GLR has spectral smoothing functionality as discussed in Section~\ref{subsubsec:GLR}, \ie, low-frequency components are better preserved. Such regularization is robust to both low density and noise in point clouds. 

\subsubsection{Inferring Data Structure via GSP-based Graph Learning}
Dong \etal \cite{dong2020graph} discuss that GSP-based graph learning frameworks enhance the model interpretability by inferring hidden relational structure from data, which leads to a better understanding of a complex system. 
In particular, GSP-based graph-learning has the unique advantage of enforcing certain desirable representations of the signals via frequency-domain analysis and filtering operations on graphs.  
For instance, models based on assumptions such as the smoothness or the diffusion of the graph signals show their superiority on geometric structure \cite{chen2019deep,wang2019dgcnn,gao2019optimized, gao2020exploring,tang2021icassp}. 
Please refer to \cite{dong2020graph} for more discussions.

\subsubsection{Monitoring Intermediate Representations via GSP}
Gripon \etal \cite{gripon2018inside} improve interpretability by using GSP to monitor the
intermediate representations obtained in a deep neural network. They demonstrate that the smoothness of the label
signal on a k-nearest neighbor feature graph is a good measure of separation of classes in these intermediate representations.

\subsection{Enhancing Robustness and Generalizability with GSP}
\subsubsection{Robustness}
"Robustness" of a deep learning network may refer to 1) robustness to noisy data or labels; 2) robustness to incomplete data; 3) robustness to a few training samples with supervised information (\eg, few-shot learning), etc. 
As discussed in Section~\ref{subsubsec:domain_knowledge}, introducing domain knowledge via GSP-based regularization would lead to geometric deep learning that is robust to noisy data and incomplete data. 
Besides, Ziko \etal \cite{ziko2020laplacian} proposed a transductive Laplacian-regularized inference for few-shot learning tasks, which encourages nearby query samples to have consistent label assignments and thus leads to robust performance. 

\subsubsection{Generalizability}
The generalizability of a model expresses how well the model will perform on unseen data. Regarding the generalizability, we assume a hypothesis: even when the unseen data may demonstrate rather different distribution characteristics, there may be an intrinsic structure embedded in the data. Such intrinsic structure usually can be better maintained from seen datasets to unseen data. That is, the data structure is assumed to be more stable than the data themselves. Consequently, when GSP tools are incorporated into deep learning networks, the graph structure (motivated by the data structure) could provide extra insights / guidance from the structure domain, in addition to the data domain, that finally enhances the generalizability of the network. For example, deep GLR \cite{zeng19cvprw} integrates graph Laplacian regularization as a trainable module into a deep learning framework, which exhibits strong cross-domain generalization ability.

\vspace{-0.05in}
\section{Future Directions}
\label{sec:future}
Regardless of the great success of GSP methods in various applications involving geometric data processing and analysis, there remain quite a few challenges ahead.
Some open problems and potential future research directions include:
\begin{itemize}

    \item GSP for time-varying geometric data processing: 
    Unlike regularly sampled videos, 4D geometric data are characterized by irregularly sampled points, both spatially and temporally, and the number of points in each time instance may also vary. This makes it challenging to establish temporal correspondences and exploit the temporal information. 
    While some works have been done in the context of 4D point cloud compression \cite{thanou2016graph, xu2019predictive} and restoration \cite{fu20icme,fu21tmm} with GSP, it still remains challenging to address complex scenarios with fast motion. 

    \item GSP for implicit geometric data processing: While not discussed in detail, the presented GSP framework embraces the processing of geometric data that is implicitly contained in the data, e.g., multi-view representations and light fields. 
    For instance, Maugey \etal \cite{maugey2015tip} proposed a graph-based representation of geometric information based on multi-view images. While this work aims at more efficient compression, we believe the use of such representations can potentially be leveraged for a wide range of inference tasks as well.

    \item GSP for enhancing model interpretability: GSP paves an insightful way to interpretable geometric deep learning, which also leads to more robust and generalizable deep learning. While we have discussed the interpretability of GNNs via GSP from several aspects in Section~\ref{sec:gnn}, we believe further steps could be made for more interpretable geometric deep learning and even reasoning in artificial intelligence.

    \item GSP for model-based geometric deep learning: as mentioned in Section~\ref{sec:gnn}, model-based GSP methods lack flexibility but perform robustly for different input data; while learning-based methods (\eg, with GNN) are highly flexible but may not generalize well.
Hence, it is desirable to explicitly \emph{integrate} GSP models (\eg, the GSRs in Section~\ref{subsec:nodal_optimization}) into learning-based methods to retain the benefits of both paradigms~\cite{zeng2019deep,valsesia2020deep}.
\end{itemize}

\vspace{-0.1in}
\section{Conclusions}
\label{sec:conclusion}
We present a generic GSP framework for geometric data, from theory to applications. 
Distinguished from other graph signals, geometric data are discrete samples of continuous 3D surfaces, 
which exhibit unique characteristics such as piece-wise smoothness that can be compactly, accurately, and adaptively represented on graphs. 
Hence, graph signal processing (GSP) is naturally advantageous for the processing and analysis of geometric data, with interpretations in both the discrete domain and the continuous domain with Riemannian geometry. 
In particular, we discuss spectral-domain GSP methods and nodal-domain GSP methods, as well as their relation. 
Further, we provide the interpretability of Graph Neural Networks (GNNs) from the perspective of GSP, highlighting that the basic graph convolution operation is essentially graph spectral or nodal filtering and that representation learning of GNNs can be understood or enhanced by GSP.
We anticipate this interpretation will inspire future research on more principled GNN designs that leverage the key GSP concepts and theory. 
Finally, we discuss potential future directions and challenges in GSP for geometric data as well as GSP-based interpretable GNN designs.

\vspace{-0.05in}
\bibliographystyle{IEEEtran}

\vspace{-0.5in}

\begin{IEEEbiography}
	[{\includegraphics[width=1in,height=1.25in,clip,keepaspectratio] {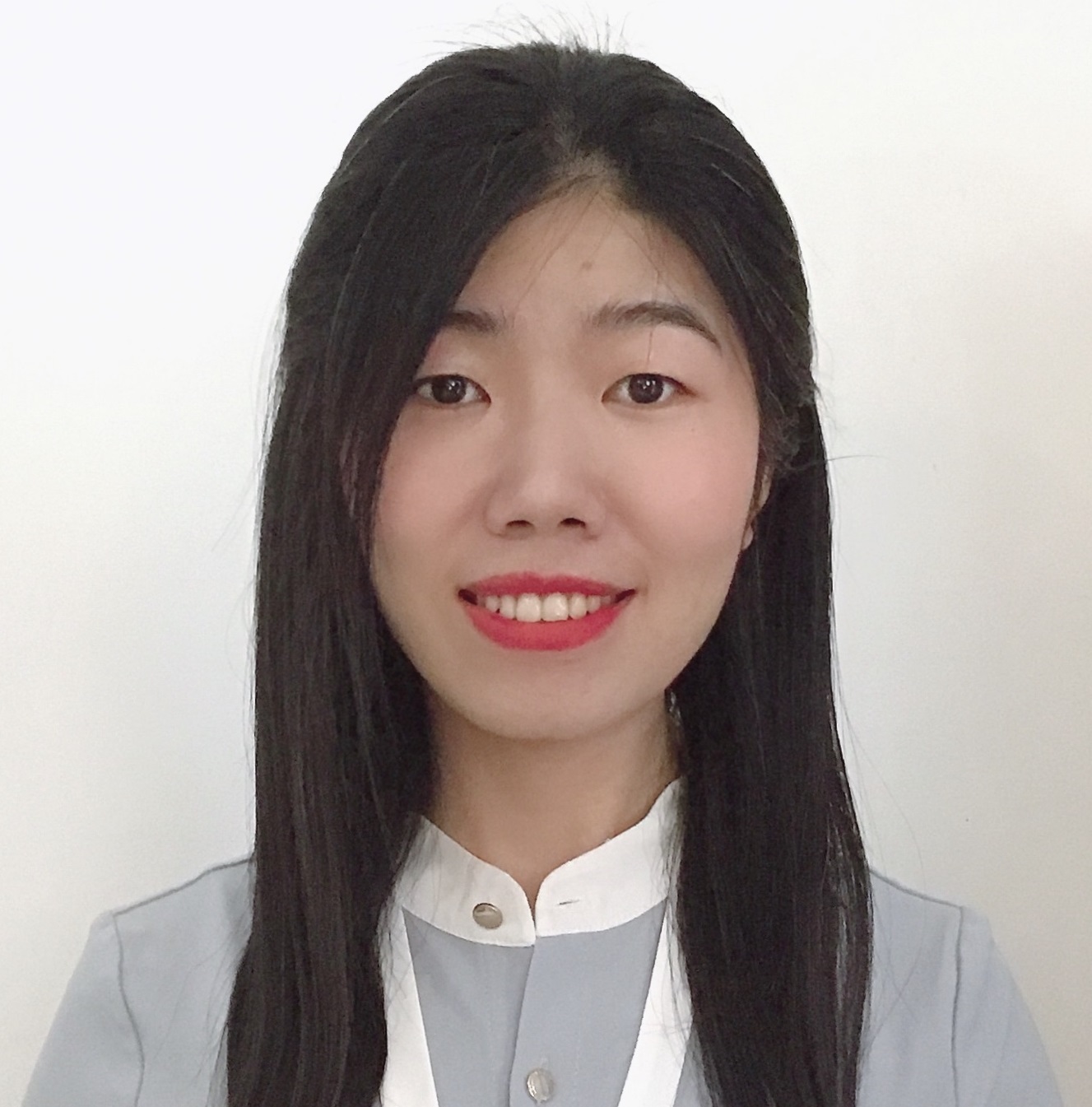}}]
	{Wei Hu}
	(Senior Member, IEEE) received the B.S. degree in Electrical Engineering from the University of Science and Technology of China in 2010, and the Ph.D. degree in Electronic and Computer Engineering from the Hong Kong University of Science and Technology in 2015.
	She was a Researcher with Technicolor, Rennes, France, from 2015 to 2017. She is currently an Assistant Professor with Wangxuan Institute of Computer Technology, Peking University. Her research interests are graph signal processing, graph-based machine learning and 3D visual computing. She has authored over 50 international journal and conference publications, with several paper awards including Best Student Paper Runner Up Award in ICME 2020 and Best Paper Candidate in CVPR 2021. 
	She was awarded the 2021 IEEE Multimedia Rising Star Award---Honorable Mention. She serves as an Associate Editor for Signal Processing Magazine, IEEE Transactions on Signal and Information Processing over Networks, etc. 
% 	She holds 4 granted U.S./Japan patents and is applying for over 10 on her signal processing techniques. 
%	She has served as a regular reviewer for IEEE Trans. on Image Processing, IEEE Trans. on Signal Processing, IEEE Trans. on Circuits and Systems for Video Technology, etc. 
\end{IEEEbiography}

\vspace{-0.5in}
\begin{IEEEbiography}
	[{\includegraphics[width=1in,height=1.25in,clip,keepaspectratio] {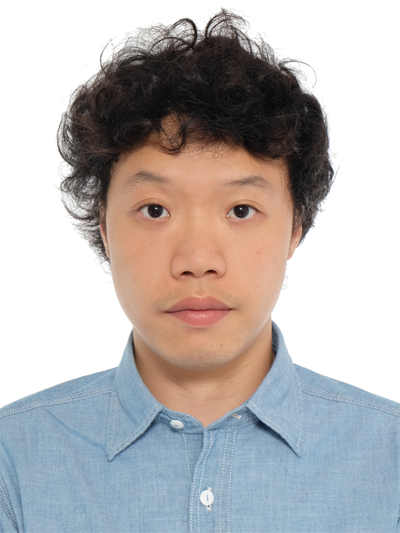}}]
	{Jiahao Pang}
	(Member, IEEE) received the B.Eng. degree from South China University of Technology, Guangzhou, China, in 2010, and the M.Sc. and Ph.D. degrees from the Hong Kong University of Science and Technology, Hong Kong, China, in 2011 and 2016, respectively. 
	He was a Senior Researcher with SenseTime Group Limited, Hong Kong, China, from 2016 to 2019.
	He is currently a Staff Engineer with InterDigital, Princeton, NJ, USA.
	His research interests include 3D computer vision, image processing, graph signal processing and deep learning.
	He has over 30 international journal and conference publications.
%	He also serves as a reviewer for IEEE TIP, IEEE CVPR, IEEE ICCV, AAAI, {\it etc}.
\end{IEEEbiography}

\vspace{-0.5in}
\begin{IEEEbiography}[{\includegraphics[width=1in,height=1.5in,clip,keepaspectratio]{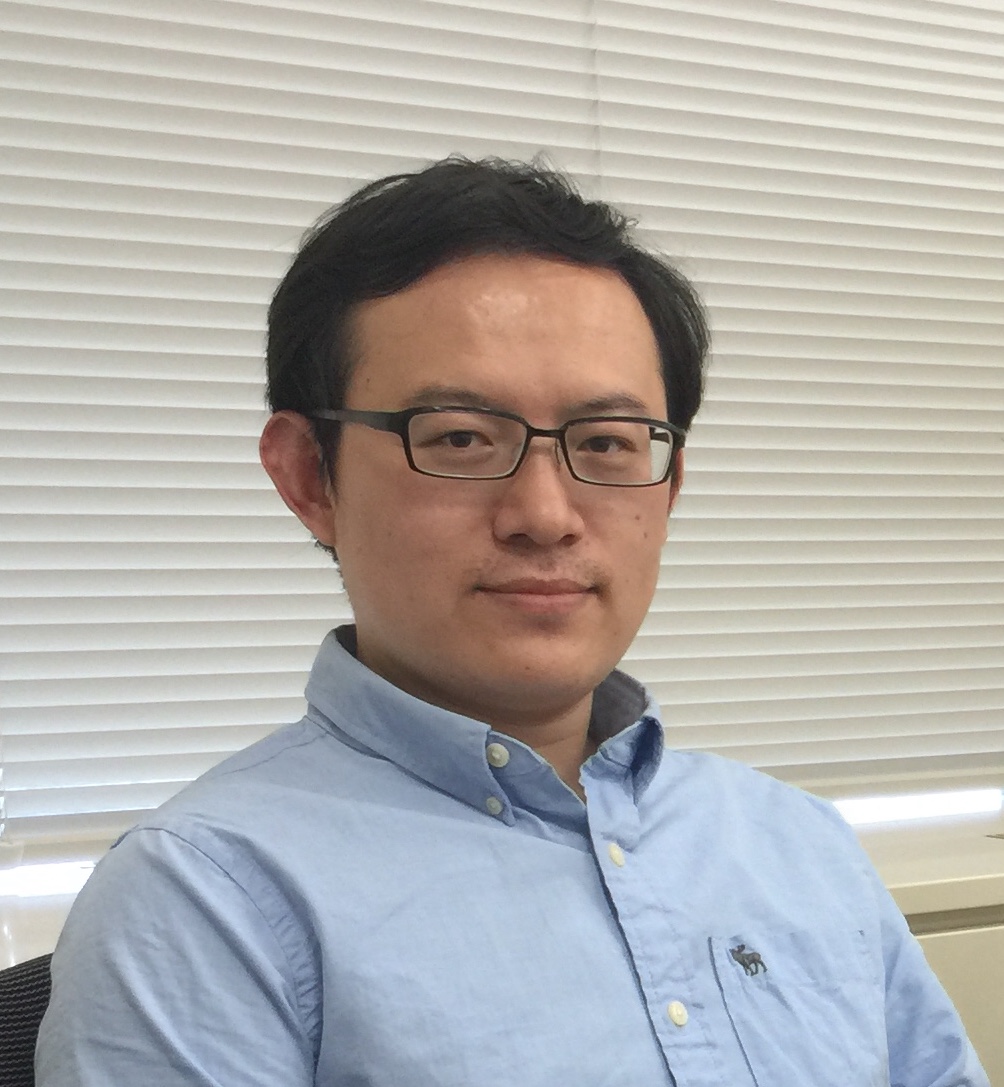}}]{Xianming Liu} (Member, IEEE)
 is a Professor with the School of Computer Science and Technology, Harbin Institute of Technology (HIT), Harbin, China. He received the B.S., M.S., and Ph.D. degrees in computer science from HIT, in 2006, 2008 and 2012, respectively. In 2011, he spent half a year at the Dept of Electrical \& Computer Engineering, McMaster University, Canada, as a visiting student, where he then worked as a post-doctoral fellow (2012-2013). He worked as a project researcher at National Institute of Informatics (NII), Tokyo, Japan (2014-2017). He has published over 80 international conference and journal publications, including top IEEE journals. %, such as T-IP, T-CSVT, T-IFS, T-MM, T-GRS, T-VCG, T-IE; and top conferences, such as CVPR, IJCAI and DCC. 
 He is the recipient of IEEE ICME 2016 Best Student Paper Award. His research interests include multimedia signal processing and computational imaging.
\end{IEEEbiography}

\vspace{-0.5in}
\begin{IEEEbiography}
	[{\includegraphics[width=1in,height=1.25in,clip,keepaspectratio] {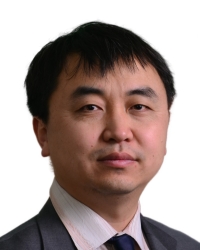}}]
	{Dong Tian}
	(Senior Member, IEEE) received the B.S. and M.Sc degrees from University of Science and Technology of China, Hefei, China, in 1995 and 1998, and the Ph.D. degree from Beijing University of Technology, Beijing, in 2001. 
	He is currently a Sr. Principal Engineer with InterDigital, Princeton, NJ, USA, after serving as a Sr. Principal Research Scientist with MERL, Cambridge, MA USA from 2010-2018, a Sr. Researcher with Thomson Corporate Research, Princeton, NJ, USA, from 2006-2010, and a Researcher with Tampere University of Technology from 2002-2005.
	His research interests include image processing, point cloud processing, graph signal processing, and deep learning.
	He has been actively contributing to both standards and academic communities. %He holds 30+ US-granted patents. He also maintains a consistent publication record on top-tier journals/transactions and conferences.
	Dr. Tian serves as an AE of TIP (2018-), General Co-Chair of MMSP'20, TPC chair of MMSP'19, etc. He is also a TC member of IEEE MMSP, and IDSP. 
\end{IEEEbiography}

\vspace{-0.5in}
\begin{IEEEbiography}
	[{\includegraphics[width=1in,height=1.25in,clip,keepaspectratio] {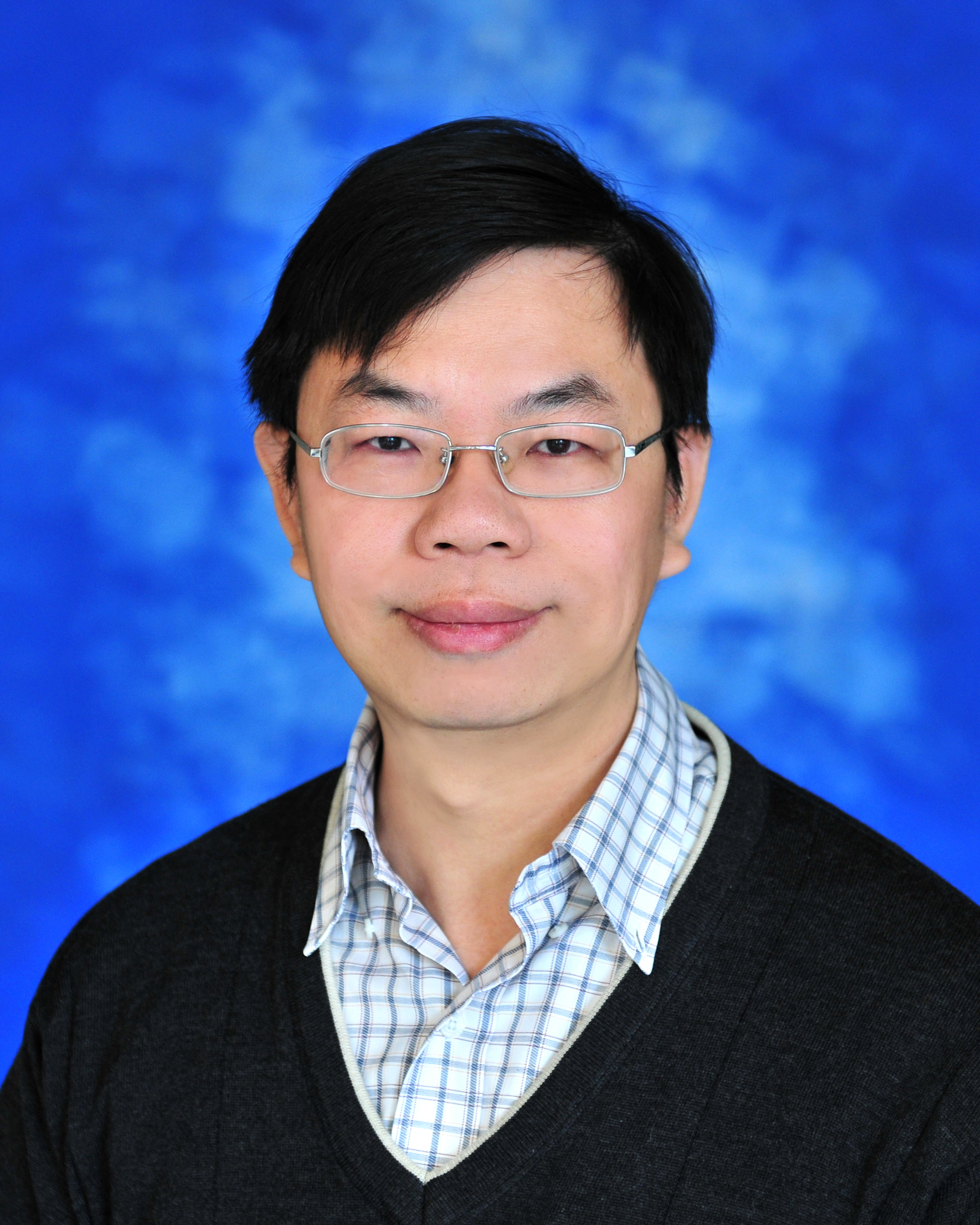}}]
	{Chia-Wen Lin}
	(Fellow, IEEE) received his Ph.D. degree from National Tsing Hua University (NTHU), Taiwan, in 2000.  
	Dr. Lin is currently Professor with the Department of Electrical Engineering and the Institute of Communications Engineering, NTHU, and R\&D Director of the Electronic and Optoelectronic System Research Laboratories, Industrial Technology Research Institute.   His research interests include image and video processing, computer vision, and video networking.  He served as Fellow evaluating Committee member (2021) and Distinguished Lecturer (2018--2019) of IEEE Circuits and Systems Society. He has served on the editorial boards of TMM, TIP, TCSVT, IEEE Multimedia, and Elsevier JVCI.  He is Chair of ICME Steering Committee and was Steering Committee member of IEEE TMM from 2013 to 2015. He served as TPC Co-Chair of ICIP 2019 and ICME 2010, and General Co-Chair of VCIP 2018. He received best paper awards from VCIP 2010 and 2015.
\end{IEEEbiography}

\vspace{-0.5in}
\begin{IEEEbiography}
	[{\includegraphics[width=1in,height=1.25in,clip,keepaspectratio] {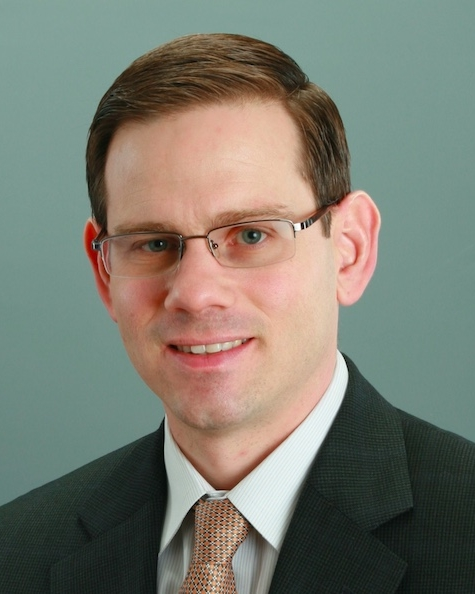}}]
	{Anthony Vetro}
    (Fellow, IEEE) received the B.S., M.S., and Ph.D. degrees in Electrical Engineering from Polytechnic University, Brooklyn, NY. Dr. Vetro is currently VP \& Director at Mitsubishi Electric Research Labs, in Cambridge, MA. He is responsible for AI related research in the areas of computer vision, speech/audio processing, and data analytics. In his 20+ years with the company, he has contributed to the development and transfer of several technologies to Mitsubishi products.
    %, including digital television receivers and displays, surveillance and camera monitoring systems, automotive equipment, as well as satellite imaging systems. 
    He has published more than 200 papers and has been a member of the MPEG and ITU-T video coding standardization committees for a number of years, serving in numerous leadership roles. 
    He is also active in various IEEE conferences, technical committees, and boards, most recently serving on the Conference Board of IEEE SPS, as a Senior AE for the Open Journal on Signal Processing, and as General Co-Chair of ICIP 2017.
    %He has served on the (senior) editorial boards of several IEEE journals: Open Journal on Signal Processing, JSTSP, JETCAS, TIP, TCSVT, SPM, and IEEE Multimedia. He was Steering Committee member of IEEE TMM and the MMSP TC Chair of IEEE SPS. He was General Co-Chair of ICIP 2017 and ICME 2015, and also served as TPC Co-Chair for ICME 2016. He received the 2003 IEEE TCSVT Best Paper Award.
\end{IEEEbiography}

\end{document}